\documentclass[%
 reprint,
 amsmath,amssymb,
 aps,
]{revtex4-2}

\usepackage[english]{babel}

\usepackage{amsmath}
\usepackage{amssymb}
\usepackage{latexsym}
\usepackage{verbatim}
\usepackage{subfigure}
\usepackage[final]{graphicx}
\usepackage{psfrag}

\newtheorem{theorem}{Theorem}
\newtheorem{proposition}{Proposition}
\newtheorem{lemma}{Lemma}

{\begin{list}               %    with flush left bullets.
    {$\bullet$ \hfill}{
        \setlength{\leftmargin}{\parindent}
        \setlength{\parsep}{0.04\baselineskip}
        \setlength{\itemsep}{0.5\parsep}
        \setlength{\labelwidth}{\leftmargin}
        \setlength{\labelsep}{0em}}
    }
{\end{list}}

\providecommand{\eref}[1]{\eqref{eq:#1}}  % call \eqref from amstex
\providecommand{\cref}[1]{Chapter~\ref{chap:#1}}
\providecommand{\sref}[1]{Section~\ref{sec:#1}}

\providecommand{\R}{\ensuremath{\mathbb{R}}}

\providecommand{\abs}[1]{\lvert#1\rvert}
\providecommand{\norm}[1]{\lVert#1\rVert}

\providecommand{\set}[1]{\left\{#1\right\}}

\providecommand{\bydef}{\overset{\text{def}}{=}}

\renewcommand{\vec}[1]{\ensuremath{\boldsymbol{#1}}}
\providecommand{\mat}[1]{\ensuremath{\boldsymbol{#1}}}

\providecommand{\mI}{\mat{I}}

\providecommand{\vh}{\vec{h}}

 \providecommand{\vs}{\vec{s}}

\providecommand{\vu}{\vec{u}}

\providecommand{\new}{\text{new}}

\providecommand{\argmax}{\text{argmax}}

\providecommand{\w}{\omega}

\usepackage{graphicx}
\usepackage{dcolumn}
\usepackage{bm}
\usepackage{hyperref}
\usepackage{microtype}
\usepackage{graphicx}
\usepackage{amssymb}
\usepackage{xcolor}
\usepackage{subfigure}
\usepackage{booktabs} 
\usepackage{placeins}

\begin{document}

\preprint{APS/123-QED}

\title{The role of regularization
in classification of high-dimensional noisy Gaussian mixture}

\author{Francesca Mignacco}
 \affiliation{Universit\'{e} Paris-Saclay, CNRS, CEA, Institut de physique th\'{e}orique, 91191, Gif-sur-Yvette, France}

\author{Florent Krzakala}
\affiliation{Laboratoire de Physique de l’Ecole normale sup\'erieure, ENS, Universit\'e PSL, CNRS, Sorbonne
Universit\'e, Universit\'e de Paris, F-75005 Paris, France}

\author{Yue M. Lu}
\affiliation{John A. Paulson School of Engineering and Applied Sciences, Harvard University, Cambridge, MA 02138, USA
}
\author{Lenka Zdeborov\'{a}}
\affiliation{Universit\'{e} Paris-Saclay, CNRS, CEA, Institut de physique th\'{e}orique, 91191, Gif-sur-Yvette, France\vspace{.5cm}
}

%\date{\today}

\begin{abstract}
We consider a high-dimensional mixture of two Gaussians in the noisy regime where even an oracle knowing the centers of the clusters misclassifies a small but finite fraction of the points. 
We provide a rigorous analysis of the generalization error of regularized convex classifiers, including ridge, hinge and logistic regression, in the high-dimensional limit where the number $n$ of samples and their dimension $d$ go to infinity while their ratio is fixed to $\alpha=n/d$.  
We discuss surprising effects of the regularization that in some cases allows to reach the Bayes-optimal performances. We also illustrate the interpolation peak at low regularization, and analyze the role of the respective sizes of the two clusters. 
\end{abstract}

\maketitle

%\tableofcontents

\section{Introduction}
\label{sec:intro}

High-dimensional statistics where both the dimensionality~$d$, and number of samples~$n$ are large with a fixed ratio $\alpha = n/d$ has largely non-intuitive behaviour. A number of the associated statistical surprises are for example presented in the recent, yet already rather influential papers \cite{hastie2019surprises,sur2019modern} that analyze high-dimensional regression for rather simple models of data. The present paper subscribes to this line of work and studies high-dimensional classification in one of the simplest models considered in statistics --- the mixture of two Gaussian clusters in $d$-dimensions, one of size $\rho n$ and the other $(1-\rho)n$ points. The labels reflect the memberships in the clusters. In particular, there are two centroids localized at $\pm \frac{{\bf v^*}}{\sqrt{d}}\in \mathbb{R}^d$, and we are given data points ${\bf x}_i, i=1\ldots n$ generated as
\begin{equation}
{\bf x}_i = \frac {\bf v^*}{\sqrt d}y_i + \sqrt{\Delta} {\bf z}_i ,
\label{eq:model}
\end{equation} 
where both ${\bf z}_i$ and ${\bf v^*}$ have components taken in ${\cal N}(0,1)$. The labels $y_i \in {\pm 1}$ are generated randomly with a fraction $\rho$ of $+1$ (and $1-\rho$ of $-1$).  We focus on the high-dimensional limit where $n,d\!\to\!\infty$ while $\alpha=n/d$, $\rho$ and $\Delta$ are fixed. The factor $\sqrt{d}$ in \eqref{eq:model} is such that a classification better than random is possible, yet even the oracle-classifier that knows exactly the centroid $\frac{{\bf v}^*}{\sqrt{d}}$ only achieves a classification error bounded away from zero.  We focus on ridge regularized learning performed by the empirical risk minimization of the loss:
\begin{eqnarray}
{\cal L}({\bf w},b)   &=&  \sum_{i=1}^n \ell\left[y_i (\tfrac{1}{\sqrt{d}}{\bf x}_i ^\top {\bf w}+b )\right]+ \frac 1{2} \lambda \|{\bf w} \|_2^2,\label{eq:loss}
\end{eqnarray} 
where ${\bf w}$ and $b$ are, respectively, the weight vector and the bias to be learned, and $\lambda$ is the tunable strength of the regularization.
While our result holds for any convex loss function $\ell(.)$, we will mainly concentrate on the following classic ones: the square $\ell(v)=\tfrac{1}{2}(1-v)^2$, the logistic $\ell(v)=\log\left(1+e^{-v}\right)$, and the hinge $\ell(v)=\max_v\{0,1-v\}$.  We shall also study the Bayes-optimal estimator, i.e. the one achieving the lowest possible test error on classification given the $n$ samples $y_i,{\bf x}_i$ and the model, including the constants $\rho$ and $\Delta$.
%, as well as a special plug-in estimator provided by the Hebbs rule \cite{hebb2005organization}. 
Crucially, the position of the centroid is {\it not known} and can only be estimated from the data.   

\paragraph*{\textbf{Our contributions and related works ---} }
%The unsupervised version of the model have also been a subject of interest \cite{baik2005phase,donoho2018optimal,lesieur2016phase,miolane2017fundamental,alaoui2018detection}.
The unsupervised version of the problem is the standard Gaussian mixture modeling problem in statistics \cite{friedman2001elements}. For the supervised model considered here, \cite{lelarge2019asymptotic} recently computed rigorously the  performance of the Bayes-optimal estimator (that knows the generative model of the data, but does not have access to the vector ${\bf v}^*$) for the case of equally sized clusters. We generalize these results for arbitrary cluster sizes  to provide a baseline for the estimators obtained by empirical risk minimization.

The model was recently under investigation in a number of papers. In \cite{mai2019high}, the authors study the same data generative model in the particular case of equally sized clusters, and analyze {\it non-regularized losses} under the assumption that the data are not linearly separable. They conclude that in that case the square loss is a universally optimal loss function. Our study of the regularized losses shows that the performance of the non-regularized square loss can be easily, and drastically improved. \cite{deng2019model} studied the logistic loss, again without regularization and for two clusters of equal size, and derive the linear separability condition in this case.

As a first contribution, we provide rigorous closed-form asymptotic formulas for the generalization and training error in the noisy high-dimensional regime, for any convex loss $\ell(.)$, that include the effects of regularization, and for arbitrary cluster size. Our proof technique uses  Gordon's inequality technique \cite{Gordon:85,Gordon:1988lr,chris:152}, as in \cite{deng2019model}. We show through numerical simulations that the formulas are extremely accurate even at moderately small dimensions.

Secondly, we present a systematic investigation of the effects of regularization and of the cluster size, discussing in particular how far estimators obtained by empirical risk minimization fall short of Bayes-optimal one, with surprising conclusions where we illustrate the effect of strong and weak regularizations. In particular, when  data are linearly separable, ~\citet{rosset2004margin} proves that all monotone non-increasing loss functions depending on the margin find a solution maximizing the margin. This is indeed exemplified in our model by the fact that for $\alpha<\alpha^*(\Delta,\rho)$ (the location of transition for linear separability) the hinge, and logistic losses converge to the same test error as the regularization tends to zero. This is related to the implicit regularization of gradient descent for the non-regularized minimization \cite{soudry2018implicit}, and we discuss this in connection with the ``double-descent" phenomenon that is currently the subject of intense studies \cite{geiger2019jamming,belkin2019reconciling,hastie2019surprises,mitra2019understanding,mei2019generalization}. 

The existence of a sharp transition for perfect separability in the model, with and without bias, is interesting in itself. Recently  \cite{candes2018phase} analyzed the maximum likelihood estimate (MLE) in high-dimensional logistic regression. While they analyzed Gaussian data (whereas we study Gaussian mixture) their results on the existence of the MLE being related to the separability of the data and having a sharp phase transition are of the same nature as ours, and similar to earlier works in statistical physics \cite{gardner1988space,gardner1989three,krauth1989storage}. 

Finally, we note that the formulas proven here can also be obtained from the heuristic replica theory from statistical physics. Indeed, a model closely related to ours was studied in this literature \cite{del1989perceptron,franz1990prosopagnosia} and our rigorous solution thus provides a further example of a rigorous proof of a result obtained by this  technique.

All these results show that the Gaussian mixtures model studied here allows to discuss, illustrate, and clarify in a unified fashion many phenomena that are currently the subject of intense scrutiny in high-dimensional statistics and machine learning.

\section{Main theoretical results}
\label{sec:main}

\subsection{Performance of empirical risk minimization}
Our first result is a rigorous analytical formula for the
generalization classification error obtained by the empirical risk minimization of \eref{loss}. Define $q$ as the length of the vector $\bf w$ and $m$ as its overlap with ${\bf v^*}$, both rescaled by the dimensionality $d$ 
\begin{equation}
    q \equiv \frac 1d \|{\bf w}\|_2^2, ~~~ m \equiv \frac 1d {\bf v^*}^\top {\bf w}\label{eq:overlaps},
\end{equation}
then we have the following:
\begin{theorem}[Asymptotics of $q$ and $m$] In the high dimensional limit when $n,d \to \infty$ with a fixed ratio $\alpha=n/d$, the length $q$ and overlap $m$ of the vector $ {\bf w}$ obtained by the empirical risk minimization of \eref{loss} with a convex loss converge to deterministic quantities given by the unique fixed point of the system:
\begin{align}
m &=& \frac {\hat m}{\lambda+\hat{\gamma}}, \label{eq:1} \\
q &=& \frac {\hat q + \hat m^2}{(\lambda+\hat{\gamma})^2},\\
\gamma &=& \frac {\Delta}{\lambda+\hat{\gamma}},\label{eq:gamma}\\
\hat m &=& \frac{\alpha}{\gamma} {\mathbb E}_{y,h}\left[v(y,h,\gamma)-h\right],\\
\hat q &=& \frac{\alpha \Delta}{\gamma^2} {\mathbb E}_{y,h}\left[(v(y,h,\gamma)-h)^2\right],\\
\hat{\gamma} &=& \frac{\alpha \Delta}{\gamma} \left(1 - {\mathbb E}_{y,h}\left[ \partial_{h} v(y,h,\gamma)\right]\right) \label{eq:6},
\end{align}
where $h \sim {\cal N}(m+yb,\Delta q)$, $\rho \in (0, 1)$ is the probability with which $y_i = 1$, and $v$ is the solution of
\begin{equation}
   v \equiv \arg\,\underset{\omega}{{\min}} \frac{(\omega-h(y,m,q,b))^2}{2\gamma} + \ell(\omega),
   \label{eq:thresholding}
\end{equation}
and the bias $b$, defined in \eref{loss}, is  the solution of the equation
\begin{equation}
  \mathbb{E}_{y,h}\left[y(v-h)\right]=0 \label{eq:11}.
\end{equation}
\label{th1}
\end{theorem}
This is proven in the next section using Gordon's minimax approach. Once the fixed point values of the overlap $m$ and length $q$ are known, then we can express the asymptotic values for the generalization error and the training loss:
\begin{theorem}[Generalization and training error] In the same limit as in theorem \ref{th1}, the generalization error expressed as fraction of wrong labeled instances is given by
\begin{equation}
\varepsilon_\text{gen} = \rho Q\Big(\frac{m + b}{ \sqrt{\Delta q}}\Big) + (1-\rho) Q\Big(\frac{m - b}{ \sqrt{\Delta q}}\Big),\label{eq:generalization}
\end{equation}
where $Q(x) = \frac{1}{\sqrt{2\pi}} \int_x^\infty e^{-t^2/2} dt$ is the Gaussian tail function.
The value of the training loss rescaled by the data dimension reads
\begin{eqnarray}\label{eq:training_error}
   {\rm L}_{\rm train} \equiv  \lim_{d \to \infty} \frac {\cal L}d
 =    
   \frac {\lambda q}{2} + \alpha\mathbb{E}_{y,h} \left[l(v(y,h,\gamma))\right]\, .
\end{eqnarray}
\label{th2}
\end{theorem}
The details on \eref{generalization} and \eref{training_error} are provided in Appendices \ref{app:generalization} and \ref{app:train}.

\subsection{MLE and Bayes-optimal estimator}

The maximum likelihood estimation (MLE) for the considered model corresponds to the optimization of the non-regularized logistic loss. This follows directly from the Bayes formula:
\begin{equation}
\begin{split}
    \log{\rm p}(y|{\rm x}) =\log \frac{{\rm p}({\rm x}|y)   {\rm p}_y(y)}{\underset{{y=\pm 1}}{\sum} {\rm p}({\rm x}|y)   {\rm p}_y(y)}\\=-\log\left(1+\exp(-c)\right),
    \end{split}
\end{equation}
where $c =\frac{2}{\Delta} y\left(\tfrac{1}{\sqrt{d}}{\rm v}^\top {\rm x}+\frac{\Delta}{2} \log \frac{\rho}{1-\rho}\right)$, therefore a simple redefinition of the variables leads to the logistic cost function that turns out to be the MLE (or rather the maximum a posteriori estimator if one allows the learning of a bias to account for the possibility of different cluster sizes).

The Bayes-optimal estimator is the ``best" possible one in the sense that it minimizes the number of errors for new labels. It can be computed as
\begin{equation}
    \hat{y}_{\new} = \arg\,\underset{y \in \pm 1}{\max}\log {\rm p}\left(y | \{\bf X, \bf y\}, \bf x_{\new}\right),
    \label{eq:yhatBO}
\end{equation}
where $\{\bf X, \bf y\}$ is the training set and $\bf x_{\new}$ is a previously unseen data point. In the Bayes-optimal setting, the model generating the data \eref{model} and the prior distributions $ {\rm p}_y$, $ {\rm p}_{\bf z}$, $ {\rm p}_{\bf v^*}$ are known. Therefore, we can compute the posterior distribution in \eref{yhatBO}:
\begin{equation}
 {\rm p}\left(y_{\new}|{\bf x_{\new}},{\bf X},{\bf y}\right)= \mathbb{E}_{{\bf v}|{\bf X},{\bf y}}\left[ {\rm p}\left(y_{\new}|{\bf x_{\new}},{\bf v}\right)\right],
\end{equation}
and applying Bayes theorem
\begin{eqnarray}
%\begin{split}
&&  {\rm p}\left(y_{\new}|{\bf x}_{\new},{\bf v}\right)\propto {\rm p} \left({\bf x}_{\new} | y_{\new},{\bf v}\right) {\rm p}_y\left(y_{\new}\right)\\
&& \propto \exp\left(
-\frac{1}{2\Delta}\sum_{i=1}^d
\left(x_{\new}^i-
\frac{y_{\new} {\rm v}^i}{\sqrt{d}}
\right)^2
\right) {\rm p}_y(y_{\new}). \nonumber
%\end{split}
\end{eqnarray}
Hence, we can compute the Bayes-optimal generalization error using
\begin{equation}
    \varepsilon_{\text{gen}}=\mathbb{P} \left(\hat{y}_{\new}\neq y_{\new}\right).
\end{equation}
This computation yields
\begin{equation}
    \varepsilon_{\rm{gen}}^{\rm BO} =\rho Q\Big(\frac{m_{\rm BO} + b_{\rm BO}}{ \sqrt{\Delta q_{\rm BO}}}\Big) + (1-\rho) Q\Big(\frac{m_{\rm BO} - b_{\rm BO}}{ \sqrt{\Delta q_{\rm BO}}}\Big),
    \label{eq:Bayes}
\end{equation}
where $m_{BO}=q_{\rm BO}=\tfrac{\alpha}{\Delta+\alpha}$ and $b_{BO}=\tfrac{\Delta}{2}\log \tfrac{\rho}{1-\rho}$. This formula is derived in the Appendix~\ref{app:BO}. The case $\rho=1/2$ was also discussed in \cite{dobriban2018high,lelarge2019asymptotic}.

Finally, it turns out that in this problem, one can reach the performances of the Bayes-optimal estimator, usually difficult to compute, efficiently using a simple plug-in estimator akin to applying the Hebb's rule \cite{hebb2005organization}. Consider indeed the weight vector averaged over the training samples, each multiplied by its label and rescaled by $\sqrt{d}$
\begin{equation}
    {\bf \hat w}_{\rm Hebb} = \frac{\sqrt{d}}{n}\sum_{\mu=1}^n y_{\mu}{\bf x}_{\mu}.
    \label{eq:plugin_w}
\end{equation}
It is straightforward to check that, for ${\bf \hat w}_{\rm Hebb}$, one has in large dimension $m=1$ and $q=(1+\tfrac{\Delta}{\alpha})$. If one further optimizes the bias (for instance by cross validation) and uses its optimal value $b=\tfrac{\Delta q}{2m}\log\tfrac{\rho}{1-\rho}$, plugging these in eq.~\eref{generalization} one reaches Bayes-optimal performance $\varepsilon_{\rm{gen}}^{\rm Hebb} = \varepsilon_{\rm{gen}}^{\rm BO}$. Since there exists a plug-in estimator that reaches the Bayes-optimal performance, it is particularly interesting to see how the ones obtained by empirical risk minimization compare with the optimal results.

\subsection{High-Dimensional Landscapes of Training Loss}

Our analysis also leads to an analytical characterization of the high-dimensional landscapes of the training loss. First, we let
\begin{equation}\label{eq:tl_mq}
\begin{aligned}
{\cal L}_\lambda(q, m, b) \bydef &\underset{{\bf w}}{\min}\, \frac{1}{d}\sum_{i=1}^{n} \ell[y_i (\tfrac{1}{\sqrt{d}} {\bf x}_i^\top {\bf w} + b)] + \frac{\lambda}{2d} \norm{{\bf w}}^2\\
&\text{subject to } \  q = \frac{1}{d}\norm{{\bf w}}^2  \text{ and } m = \frac{1}{d}{\bf w}^\top {\bf v}^\ast 
\end{aligned}
\end{equation}
to denote the normalized training loss when we restrict the weight vector to satisfy the two conditions in \eqref{eq:tl_mq}. In what follows, we refer to ${\cal L}_\lambda(q, m, b)$ as the ``local training loss'' at fixed values of $q, m$ and $b$. The ``global training loss'' can then be obtained as
\begin{equation}\label{eq:tl}
{\cal L}^\ast_\lambda \bydef \underset{m^2 \le q, b}{\min} {\cal L}_\lambda(q, m, b),
\end{equation}
where the constraint that $m^2 \le q$ is due to the Cauchy-Schwartz inequality: $\abs{m} = \frac{\abs{{\bf w}^\top {\bf v}^\ast}}{d} \le \frac{\norm{{\bf w}}}{\sqrt{d}} \frac{\norm{{\bf v}^\ast}}{\sqrt{d}} = \sqrt{q}$. 

In the high-dimensional limit when $n, d \to \infty$ with a fixed ratio $\alpha = n/d$, many properties of the local training loss can be characterized by a deterministic function, defined as
\begin{equation}\label{eq:E_det}
{\cal E}_\lambda(q, m, b) \bydef \alpha \mathbb{E}[\ell(v_{\gamma^\ast})] + \frac{\lambda q}{2}.
\end{equation}
Here, for any $\gamma \ge 0$, $v_\gamma$ denotes a random variable whose cumulative distribution function is given by
\begin{equation}\label{eq:v_CDF}
\begin{aligned}
\mathbb{P}(v_r \le v) &= \rho Q\left(\frac{\gamma \ell'(v) + v - m - b}{\sqrt{\Delta q}}\right)\\
&\qquad+ (1-\rho)Q\left(\frac{\gamma \ell'(v) + v - m + b}{\sqrt{\Delta q}}\right).
\end{aligned}
\end{equation}
Moreover, $\gamma^\ast$ in \eqref{eq:E_det} is the unique solution to the equation
\begin{equation}\label{eq:fix_point_gamma}
\alpha \gamma^2 \mathbb{E}[ (\ell'(v_\gamma))^2] = \Delta (q- m^2).
\end{equation}

\begin{proposition}\label{prop:Gordon_det}
Let $\Omega$ be an arbitrary compact subset of $\set{(q, m, b): m^2 \le q}$. We define
\[
{\cal L}_\lambda(\Omega) = \inf_{(q, m, b) \in \Omega} {\cal L}_\lambda(q, m, b)
\]
and
\[
{\cal E}_\lambda(\Omega) = \inf_{(q, m, b) \in \Omega} {\cal E}_\lambda(q, m, b).
\]
For any constant $\delta > 0$ and as $n, d\to \infty$ with $\alpha =n/d$ fixed, it holds that
\begin{equation}\label{eq:comp_phi_e}
\mathbb{P}\Big({\cal L}_\lambda(\Omega) \ge {\cal E}_\lambda(\Omega) - \delta\Big) \to 1.
\end{equation}
Moreover,
\begin{equation}\label{eq:comp_phi_e_ast}
{\cal L}_\lambda^\ast \to {\cal E}_\lambda^\ast \bydef \inf_{m^2 \le q, b} {\cal E}_\lambda(q, m, b),
\end{equation}
where ${\cal L}_\lambda^\ast$ is the global training loss defined in \eqref{eq:tl}.
\end{proposition}

The characterization in \eref{comp_phi_e_ast} shows that the global training loss will concentrate around the fixed value ${\cal E}_\lambda^\ast$. Meanwhile, \eref{comp_phi_e} implies that the deterministic function ${\cal E}_\lambda(q, m, b)$ serves as a high-probability \emph{lower bound} of the local training loss ${\cal L}_\lambda(\Omega)$ over any given compact subset $\Omega$. This latter property allows us to study the high-dimensional landscapes of the training loss as we move along the 3-dimensional space of the parameters $q, m$ and $b$. 

In particular, by studying ${\cal E}_\lambda(q, m, b)$, we can obtain the phase transition boundary characterizing the critical value of $\alpha$ below which the training data become perfectly separable.

\begin{proposition}\label{prop:phase_transition}
Let $\lambda = 0$. Then
\[
{\cal E}_\lambda^\ast = \begin{cases}
> 0, & \text{if } \alpha > \alpha^*\\
0, & \text{if } \alpha < \alpha^*,
\end{cases}
\]
where
\begin{align}
\alpha^* &\bydef \max_{0 \le r \le 1, b} \eta(r, b) \\
\eta(r, b) &\!=\! \frac{1-r^2}{ \int_0^\infty \!\! u^2 [\rho f(u + \frac{r}{\sqrt{\Delta}} - b) \!+ \! (1-\rho) f(u + \frac{r}{\sqrt{\Delta}}+b)] du} \nonumber 
\end{align}
and $f(x)$ is the probability density function of ${\cal N}(0,1)$.
\end{proposition}

\section{Proof Sketches}
\label{sec:sketches}

In this section, we sketch the proof steps behind our main results presented in \sref{main}. The full technical details are given in the Appendix~\ref{app:train}. 

Roughly speaking, our proof strategy consists of three main ingredients: (1) Using Gordon's minimax inequalities \cite{Gordon:85,Gordon:1988lr,chris:152}, we can show that the random optimization problem associated with the local training loss in \eref{tl_mq} can be compared against a much simpler optimization problem (see \eqref{eq:E_vec} in \sref{Gordon}) that is essentially decoupled over its coordinates; (2) we show in \sref{Gordon_det} that the aforementioned simpler problem concentrates around a well-defined deterministic limit as $n, d \to \infty$; and (3) by studying properties of the deterministic function, we reach the various characterizations given in Theorem~\ref{th1}, Proposition~\ref{prop:Gordon_det} and Proposition~\ref{prop:phase_transition}.

\subsection{The dual formulation and Gordon's inequalities}
\label{sec:Gordon}

The central object in our analysis is the local training loss ${\cal L}_\lambda(q, m, b)$ defined in \eqref{eq:tl_mq}. The challenge in directly analyzing \eqref{eq:tl_mq} lies in the fact that it involves a $d$-dimensional (random) optimization problem where all the coordinates of the weight vector $\bf w$ are fully coupled. Fortunately, we can bypass this challenge via Gordon's inequalities, which allow us to characterize ${\cal L}_\lambda(q, m, b)$ by studying a much simpler problem. To that end, we first need to rewrite \eref{tl_mq} as a minimax problem, via a Legendre transformation of the convex loss function $\ell(v)$:
\begin{equation}\label{eq:Legendre}
\ell(v) = \max_u \set{v u - \widetilde{\ell}(u)},
\end{equation}
where $\widetilde{\ell}(u)$ is the convex conjugate, defined as
\[
\widetilde{\ell}(u) = \max_v \set{u v - \ell(v)}.
\]

For example, for the square, logistic, and hinge losses defined in \sref{intro}, their corresponding convex conjugates are given by
\begin{align}
\widetilde{\ell}_\text{square}(u) &= \frac{u^2}{4} + u\\
\widehat{\ell}_\text{logistic}(u) &= \begin{cases} -H(-u), &\text{for } -1 \le u \le 0\\
\infty, &\text{otherwise}
\end{cases},
\end{align}
where $H(u) \bydef -u \log u - (1-u) \log(1-u)$ is the binary entropy function, and
\[
\widehat{\ell}_\text{hinge}(u) = \begin{cases}
u, &\text{for } -1 \le u \le 0\\
\infty, &\text{otherwise},
\end{cases}
\]
respectively.

Substituting \eref{Legendre} into \eref{tl_mq} and recalling the data model \eref{model}, we can rewrite \eref{tl_mq} as the following minimax problem
\[
\begin{aligned}
&{\cal L}_\lambda(q, m, b) =\frac{\lambda q}{2} + \\
&\underset{{\bf w} \in {\cal S}_{q, m}}{\min}\!\! \max_{{\bf u}} \frac{1}{d}\sum_{i=1}^n u_i \! \left(\frac{{\bf w}^\top {\bf v}^\ast}{d} \!\! +\!\!  \sqrt{\Delta} \frac{y_i {\bf z}_i^\top {\bf w}}{\sqrt{d}} + b y_i\right) \!\! - \widehat{\ell}(u_i)\, ,
\end{aligned}
\]
where ${\cal S}_{q, m} \bydef \set{{\bf w}: q = \frac{1}{d}\norm{{\bf w}}^2  \text{ and } m = \frac{1}{d}{\bf w}^\top {\bf v}^\ast }$.

\begin{proposition}\label{prop:Gordon_E}
For every $(q, m, b)$ satisfying $q > m^2$, let 
\begin{equation}\label{eq:E_vec}
\begin{aligned}
&{\cal E}^{(d)}_\lambda(q, m, b)\bydef \frac{\lambda q}{2} \\
&+ \max_{\vu \in \R^n} \set{-\sqrt{\frac{ \Delta_d\norm{\vu}^2(q - m^2)}{d}} + \frac{\vu^\top \vh}{d} -\frac{1}{d} \sum_{i=1}^n \widetilde{\ell}(u_i)},
\end{aligned}
\end{equation}
where $\Delta_d \bydef (Q_d / d) \Delta$ with $Q_d \sim \chi^2_d$,
\begin{equation}\label{eq:h_vec}
\vh = \sqrt{\Delta q} \vec{s} +m \vec{1} + b [y_1, y_2, \ldots, y_n]^\top
\end{equation}
and $\vs \sim \mathcal{N}(0, \mI_n)$ is an i.i.d. Gaussian random vector. Then for any constant $c$ and $\delta > 0$, we have
\begin{equation}\label{eq:Gordon_E1}
\mathbb{P}({\cal L}_\lambda(q, m, b) < c) \le 2 \mathbb{P}({\cal E}^{(d)}_\lambda(q, m, b) < c)
\end{equation}
and
\begin{equation}\label{eq:Gordon_E2}
\mathbb{P}(\abs{{\cal L}_\lambda^\ast - c} > \delta) \le 2 \mathbb{P}(\abs{\inf_{q, m, b} {\cal E}^{(d)}_\lambda(q, m, b) - c} > \delta).
\end{equation}
\end{proposition}
The proof of Proposition~\ref{prop:Gordon_E}, which can be found in the Appendix~\ref{app:proof_Gordon_E}, is based on an application of Gordon's comparison inequalities for Gaussian processes \cite{Gordon:85,Gordon:1988lr,chris:152}. Similar techniques have been used by the authors of \cite{deng2019model} to study the Gaussian mixture model for the non-regularized logistic loss for two clusters of the same size.

\subsection{Asymptotic Characterizations}
\label{sec:Gordon_det}

The definition of ${\cal E}^{(d)}_\lambda(q, m, b)$ in \eref{E_vec} still involves an optimization with an $n$-dimensional vector $\vu$, but it can be simplified to a one-dimensional optimization problem with respect to a Lagrange multiplier $\gamma$:
\begin{lemma}\label{lemma:Lagrange}
\begin{equation}\label{eq:E_vec_d}
\begin{aligned}
&{\cal E}^{(d)}_\lambda(q, m, b) = \frac{\lambda q}{2}\\
&+ \max_{\gamma > 0}\Big\{-\sqrt{\frac{\Delta_d(q-m^2) \norm{\vu_\gamma}^2}{d}} + \frac{\vu_\gamma^\top h}{d} - \frac{1}{d} \sum_{i=1}^n \widetilde{\ell}(u_{\gamma,i})\Big\},
\end{aligned}
\end{equation}
where $\vu_\gamma \in \R^n$ is the solution to 
\begin{equation}\label{eq:u_vec_h}
\nabla \widetilde{\ell}(\vu_\gamma) + \gamma \vu_\gamma = \vh,
\end{equation}
with $\vh$ defined as in \eref{h_vec}.
\end{lemma}

One can show that the problem in \eref{E_vec_d} reaches its maximum at a point $\gamma^\ast$ that is the unique solution to 
\begin{equation}\label{eq:fix_point_d}
\alpha \gamma^2 \frac{\norm{\vu_\gamma}^2}{n} = \Delta_d (q- m^2).
\end{equation}
Moreover,
\begin{equation}\label{eq:E_s_d}
{\cal E}^{(d)}_\lambda(q, m, b) = \frac{\sum_{i=1}^n \big[u_{\gamma^\ast, i} \widetilde{\ell}'(u_{\gamma^\ast,i}) - \widetilde{\ell}(u_{\gamma^\ast,i})\big]}{d} + \frac{\lambda q}{2}.
\end{equation}

In the asymptotic limit, as $n, d \to \infty$, both \eref{fix_point_d} and \eref{E_s_d} converge towards their deterministic limits:
\begin{equation}\label{eq:fix_point_u}
\alpha \gamma^2 \mathbb{E}[u_\gamma^2] = \Delta_d (q- m^2)
\end{equation}
and
\begin{equation}\label{eq:E_u_d}
{\cal E}^{(d)}_\lambda(q, m, b) \to \alpha \mathbb{E}[u_{\gamma} \widetilde{\ell}'(u_{\gamma}) - \widetilde{\ell}(u_{\gamma})] + \frac{\lambda q}{2},
\end{equation}
where $u_\gamma$ is a random variable defined through the implicit equation $\widetilde{\ell}'(u_{\gamma}) + \gamma u_\gamma = h$.

Note that \eref{fix_point_u} and \eref{E_u_d} already resemble their counterparts \eref{fix_point_gamma} and \eref{E_det} given in our main results. The precise connection can be made by introducing the following scalar change of variables: $v = \widetilde{\ell}'(u)$. It is easy to verify from properties of Legendre transformations that
\[
u = \ell'(v) \quad\text{and}\quad u \widetilde{\ell}'(u) - \widetilde{\ell}(u) = \ell(v).
\]
Substituting these identities into \eref{fix_point_u} and \eref{E_u_d} then gives us the characterizations \eref{fix_point_gamma} and \eref{E_det} as stated in \sref{main}.

Finally, the fixed point characterizations given in Theorem~\ref{th1} can be obtained by taking derivatives of ${\cal E}_\lambda(q, m, b)$ with respect to $q, m, b$ and setting them to $0$. Similarly, the phase transition curve given in Proposition~\ref{prop:phase_transition} can be obtained by quantifying the conditions under which the deterministic function ${\cal E}_\lambda(q, m, b)$ reaches its minimum at a finite point. We give more details in Appendix~\ref{app:proof_phase_transition} - \ref{app:proof_mapping}.

\begin{figure*}[ht]
\includegraphics[scale=0.3]{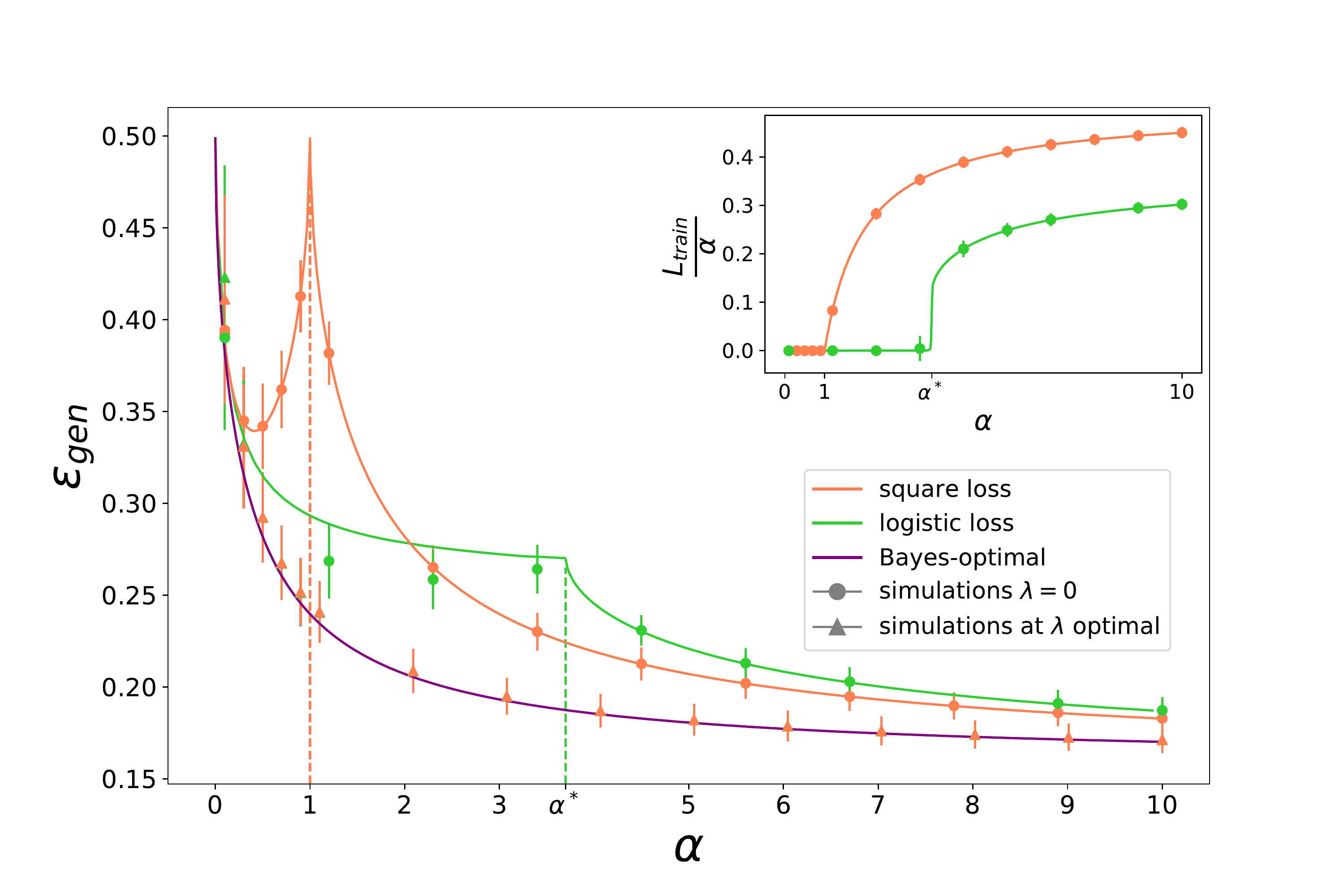}
\hspace{-0.6cm}
\includegraphics[scale=0.3]{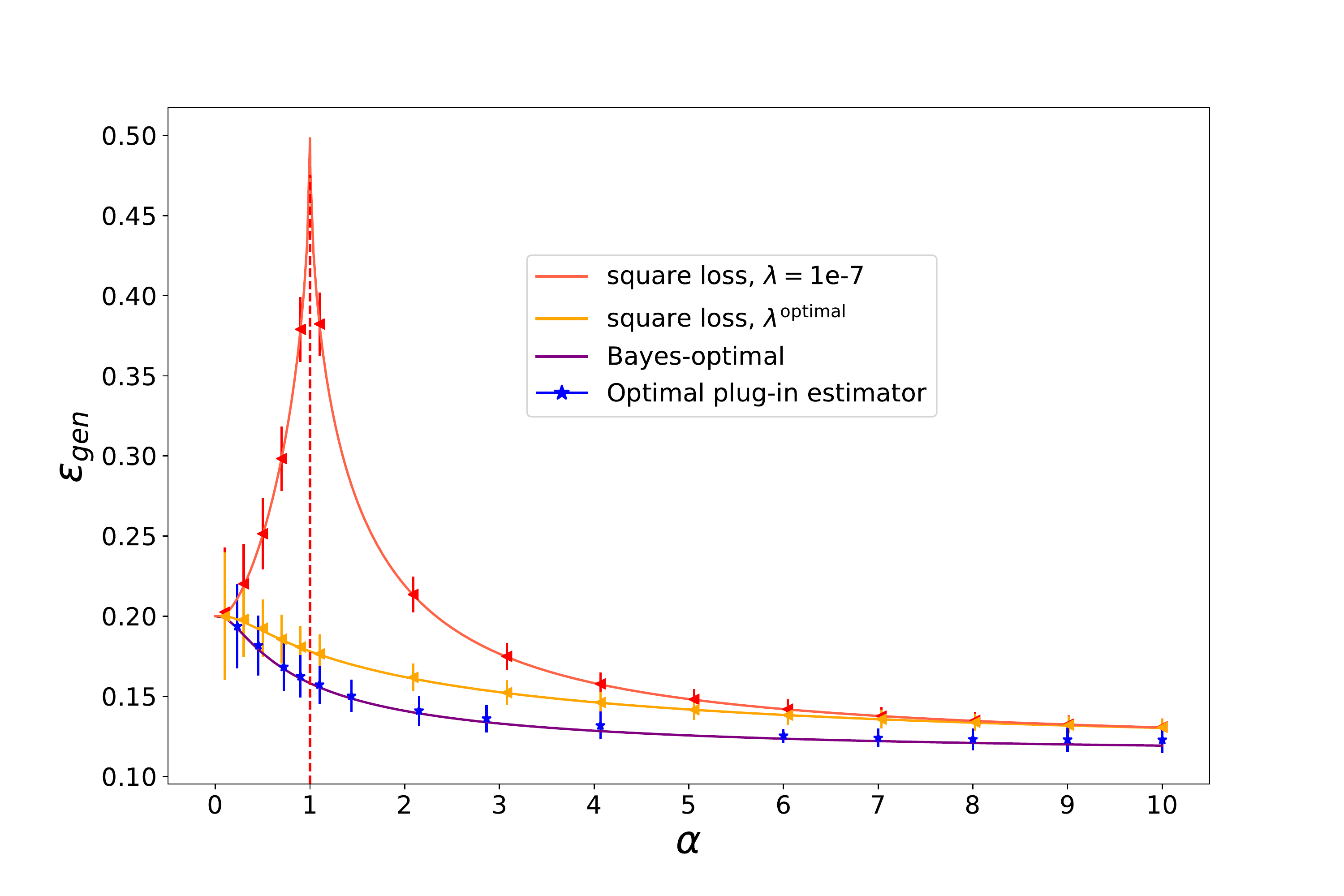}
\caption{{\bf Left (equal cluster size)}. Generalization error as a function of $\alpha$ at low regularization ($\lambda=10^{-7}$) and fixed $\Delta=1$, $\rho=0.5$. The dashed vertical lines mark the interpolation thresholds. The generalization error achieved by the square and logistic losses is compared to the Bayes-optimal one. In this symmetric clusters case, it is possible to tune $\lambda$ in order to reach the optimal performance. In the inset, the training loss as a function of $\alpha$. The training loss is close to zero up to the interpolation transition. We compare our theoretical findings with simulations, at $d=1000$. {\bf Right (unequal cluster size)} Generalization error as a function of $\alpha$ at fixed $\Delta=1$, $\rho=0.2$. The performance of the square loss at low ($\lambda=10^{-7}$) and optimal regularization is compared to the Bayes-optimal performance. In this non-symmetric case $\rho \neq 0.5$, the Bayes-optimal error is not achieved by the optimally regularized losses under consideration. We compare our results with numerical simulations at $d=1000$. Additionally, we illustrate that the Bayes-optimal performance can be reached by the optimal plug-in estimator defined in eq.~\eref{plugin_w} (here with $d=5000$).}
\label{fig:unreg_gen_vs_alpha}
\end{figure*}

\subsection{Interpretation from the replica method} 

These same equations can be independently derived from the non-rigorous replica methods from statistical physics \cite{mezard1987spin}, a technique that has proven useful in the study of high-dimensional statistical models, for instance following \cite{franz1990prosopagnosia,lesieur2016phase}. Alternatively, these equations can also be seen as a special case of the State Evolution equation of the Approximate Message Passing  algorithm \cite{donoho2009message,bayati2011dynamics,lesieur2016phase}. Both interpretations can be useful, since  the various quantities enjoy additional heuristic interpretations that allow us to obtain further insight. For instance, the parameter $\gamma$ in \eqref{eq:gamma} is  connected to the rescaled variance of the estimator ${\bf w}$:
\begin{equation}V=\lim_{d\rightarrow \infty}\frac{\mathbb{E}_{{\bf X, y}}\left[\norm{{\bf w}}^2\right]-\mathbb{E}_{{\bf X, y}}\left[\norm{{\bf w}}\right]^2}{d}.\end{equation}
The zero temperature limit of the fixed point equations obtained with the replica method corresponds to the loss minimization \cite{mezard1987spin,mezard2009information}. In this limit, the behaviour of the rescaled variance $V$ at zero penalty ($\lambda=0$) is an indicator of data separability. In the non-separable regime, the minimizer of the loss is unique and $V\rightarrow 0$ at temperature $T=0$. The parameter $\gamma$ turns out to be simply $\gamma=\tfrac{V}{T}$. However, in the regime where data are separable there is a degeneracy of solutions at $\lambda=0$, and the variance is finite: $V>0$. Hence the parameter $\gamma$ has a divergence at the transition, and this provides a very easy way to compute the location of the phase transition.

\section{Consequences of the formulas}
In this section we evaluate the above formulas and investigate how does the test error depend on the regularization parameter $\lambda$, the fraction taken by the smaller cluster $\rho$, the ratio between the number of samples and the dimension $\alpha$ and the cluster variance $\Delta$.  The details on the evaluation and iteration of the fixed point equations in Theorem \ref{th1} are provided in Appendices \ref{app:fixed_point} and \ref{app:numerics} respectively. Keeping in mind that minimization of the non-regularized logistic loss corresponds in the considered model to the maximum likelihood estimation (MLE), we thus pay a particular attention to it as a benchmark of what the most commonly used method in statistics would achieve in this problem. Another important benchmark is the Bayes-optimal performance that provides a threshold that no algorithm can improve. 

\begin{figure*}[ht]
\includegraphics[scale=0.3]{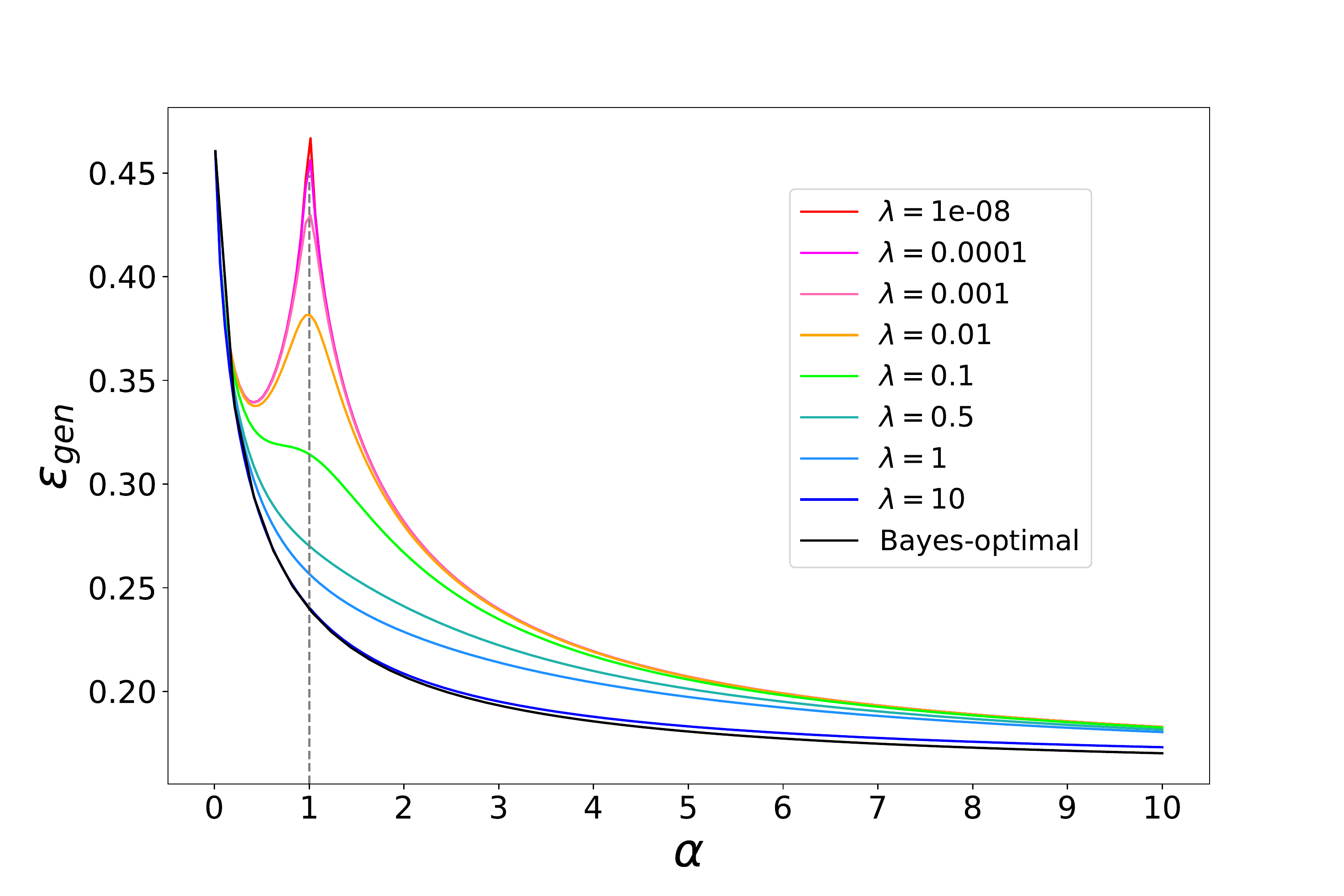}
\hspace{-0.6cm}
\includegraphics[scale=0.3]{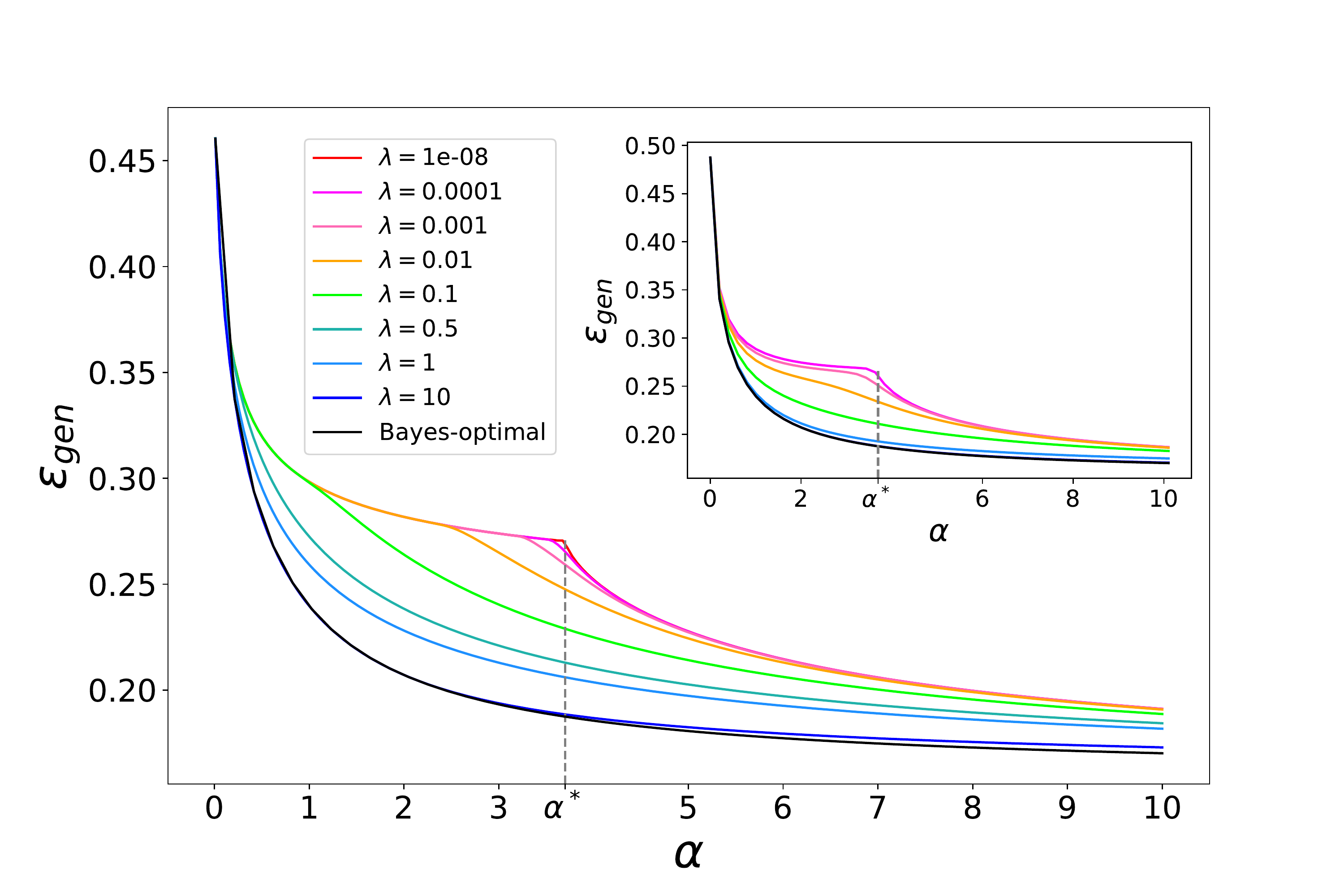}
\caption{Generalization error as a function of $\alpha$ for different values of $\lambda$, at fixed $\Delta=1$ and $\rho=0.5$, for the square loss (left), the hinge loss (right) and the logistic loss (inset), compared to the Bayes-optimal error. If the two clusters have the same size, the Bayes-optimal error can be reached by increasing the regularization. Notice how regularization smooths the curves and makes the ``peak" or ``kink" disappear in all cases.}
\label{fig:dep_L_hinge_L2}
\end{figure*}

\paragraph*{\textbf{Weak and strong regularization} ---} Fig.~\ref{fig:unreg_gen_vs_alpha} summarizes how the regularization parameter $\lambda$ and the cluster size $\rho$ influence the generalization performances. The left panel of  Fig.~\ref{fig:unreg_gen_vs_alpha} is for the symmetric case $\rho\!=\!0.5$, the right panel for the non-symmetric case $\rho=0.2$.  Let us define as $\alpha^*$  the value of $\alpha$ such that for $\alpha<\alpha^*$ the training loss for hinge and logistic goes to zero (in other words, the data are linearly separable \cite{candes2018phase}. In the left part of Fig.~\ref{fig:unreg_gen_vs_alpha} we depict (in green) the performance of the non-regularized logistic loss a.k.a. the maximum likelihood. For $\alpha > \alpha^*(\rho,\Delta)$ the training data are not linearly separable and the minimum training loss is bounded away from zero.  For $\alpha < \alpha^*(\rho,\Delta)$ the data are linearly separable, in which case properly speaking the maximum likelihood is ill-defined \cite{sur2019modern}, the curve that we depict is the limiting value reached as $\lambda \to 0^+$. The points are results of simulations with a standard scikitlearn \cite{scikit} package.
%, the agreement with the theory is highlight the implicit regularization in the simulations. 
As shown in \cite{soudry2018implicit}, even though the logistic estimator does not exist, gradient descent actually converges to the max-margin solution in this case, or equivalently to the least norm solution corresponding to $\lambda\!\to\!0^+$, a phenomenon coined ``implicit regularization", which is well illustrated here.

Another interesting phenomenon is the non-monotonicity of the curve. This is actually an avatar of the so-called ``double descent" phenomenon where the generalization ``peaks" to a bad value and then decays again. This was observed and discussed recently in several papers \cite{geiger2019jamming,belkin2019reconciling,hastie2019surprises,mitra2019understanding,mei2019generalization}, but similar observations appeared as early as 1996 in \citet{opper1996statistical}.  Indeed, we observed that the generalization error of the non-regularized square loss (in red) has a peak at $\alpha=1$ at which point the data matrix in the non-regularized square loss problem becomes invertible. It is interesting that for $\alpha > \alpha^*$ the generalization performance of the non-regularized square loss is better than the one of the maximum likelihood. This has been proven recently in \cite{mai2019high}, who showed that among all the convex {\it non-regularized losses}, the square loss is optimal. 

Fig.~\ref{fig:unreg_gen_vs_alpha} further depicts (in purple) the Bayes-optimal error eq.~(\ref{eq:Bayes}). We have also evaluated the performance of both the logistic and square loss at optimal value of the regularization parameter $\lambda$. This is where the symmetric case (left panel) differs {\it crucially} from the non-symmetric one (right panel). While in the high-dimensional limit of the symmetric case the optimal regularization $\lambda_{\rm opt}\!\to\!\infty$ and the corresponding error matches exactly the Bayes-optimal error, for the non-symmetric case $0< \lambda_{\rm opt} < \infty$ and the error for both losses is bounded away from the Bayes-optimal one for any $\alpha\!>\!0$. 

We give a fully analytic argument in the Appendix~\ref{app:BO_large_lambda} for the perhaps unexpected property of achieving the Bayes-optimal generalization at $\lambda_{\rm opt} \to \infty$ and $\rho=0.5$ for any loss that has a finite 2nd derivative at the origin.  In simulations for finite value of $d$ we use a large but finite value of $\lambda$, details on the simulation are provided in the Appendix~\ref{app:numerics}.

% Dependence on lambda

\paragraph*{\textbf{Regularization and the interpolation peak} ---} In Fig.~\ref{fig:dep_L_hinge_L2} we depict the dependence of the generalization error on the regularization $\lambda$ for the symmetric $\rho=0.5$ case for the square, hinge and logistic loss. The curves at small regularization show the interpolation peak/cusp at $\alpha=1$ for the square loss and $\alpha^*$ for all the losses that are zero whenever the data are linearly separable. We observe a smooth disappearance of the peak/cusp as regularization is added, similarly to what has been observed in other models that present the interpolation peak \cite{hastie2019surprises,mei2019generalization} in the case of the square loss. Here we thus show that a similar phenomena arises with the logistic and hinge losses as well; this is of interest as this effect has been observed in deep neural networks using a logistic/cross-entropy loss \cite{geiger2019jamming,nakkiran2019deep}. In fact, as the regularization increases, the error gets better in this model with equal-size cluster, and one reaches the Bayes-optimal values for large regularization.

\begin{figure}[ht]
    \centering
    \includegraphics[scale=0.3]{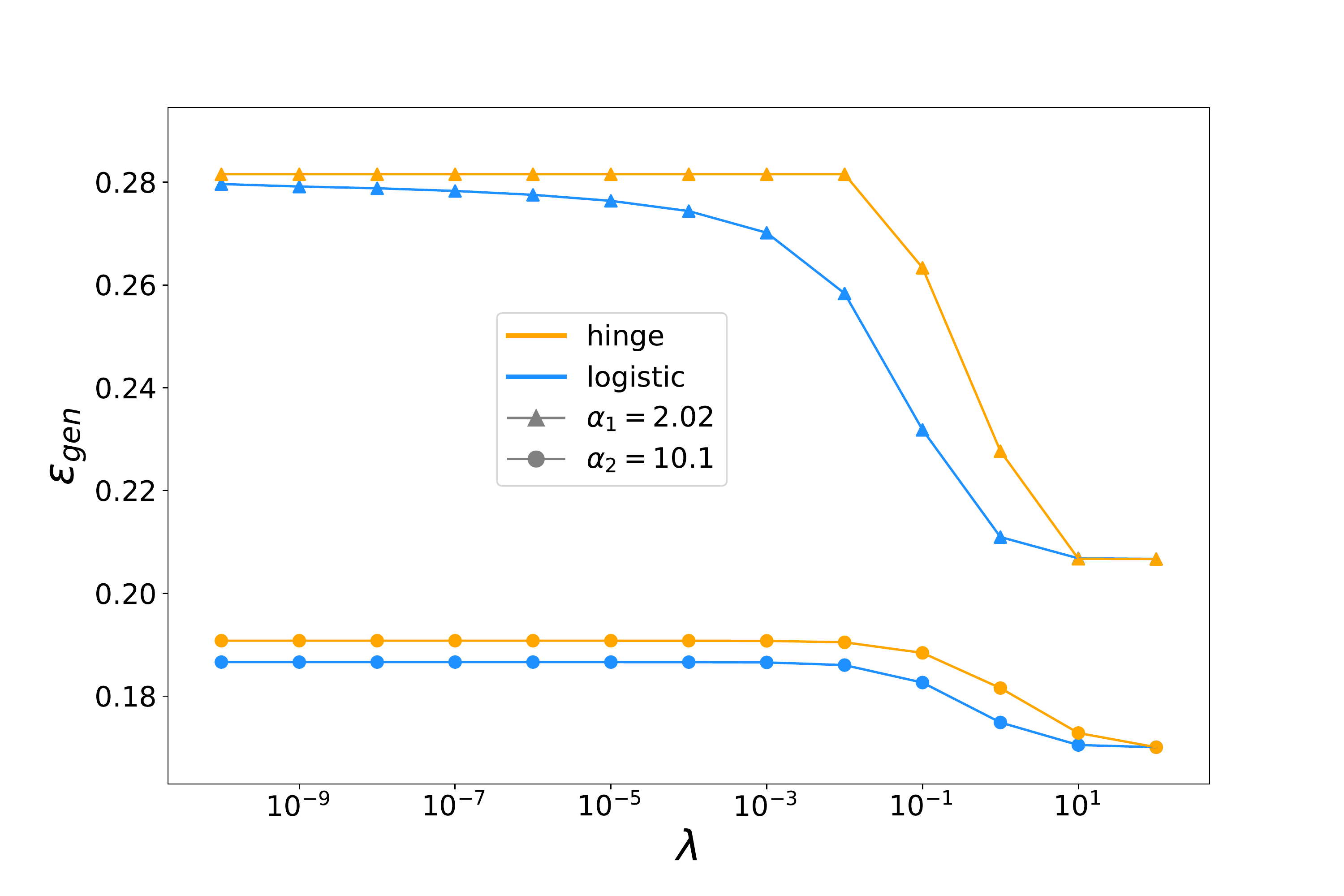}
    \vspace{-10mm}
    \caption{Generalization error as a function of $\lambda$ for the hinge and logistic losses, at fixed $\Delta=1$, $\rho=0.5$ and two different values of $\alpha$: $\alpha_1=2, \alpha_2=10$. As $\lambda \rightarrow 0^+$, the error of the two losses approaches the same value if the data are separable ($\alpha_1 <\alpha^*$). This is not true if the data are  not separable ($\alpha_2>\alpha^*$). At large $\lambda$, the error of both losses reaches the Bayes-optimal, for all $\alpha$.}
    \label{fig:log_vs_hinge}
\end{figure}

\begin{figure}[t]
\includegraphics[scale=0.35]{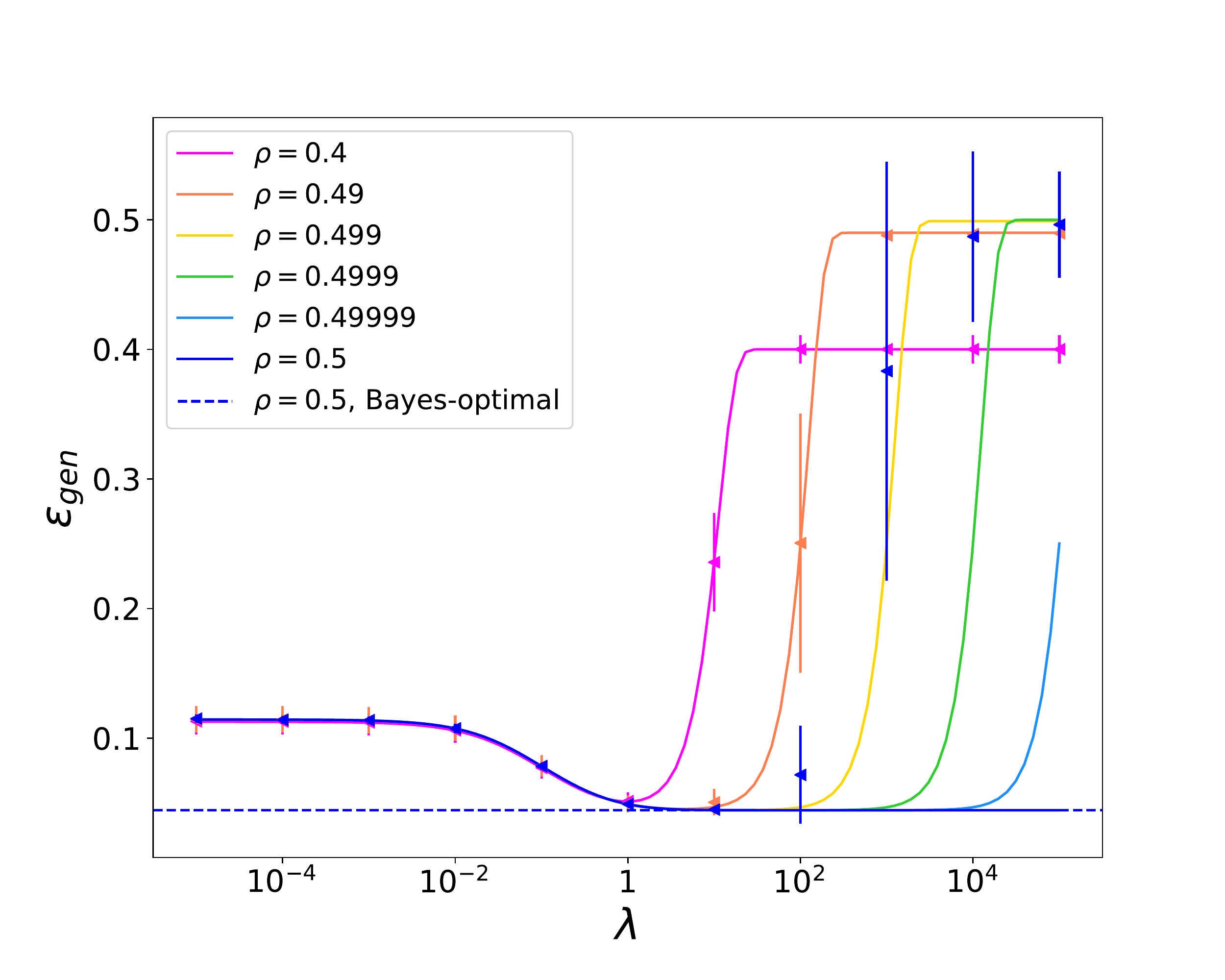}
 \vspace{-7mm}
\caption{\label{fig:gen_vs_L_L2}Generalization error as a function of $\lambda$ for different values of $\rho$ close to $0.5$, at fixed $\Delta=0.3$ and $\alpha=2$, for the square loss. At all $\rho<0.5$, the error exhibits a minimum at finite $\lambda=\lambda^*$, and reaches a plateau at $\lambda>\lambda^*$. The value of the error at the plateau is $\varepsilon_{\text{gen}}=\text{min}\{ \rho,1-\rho\}$, which is the error attained by the greedy strategy of assigning all points to the larger cluster. We compare our analytical results with simulations for $\rho=0.4,0.49,0.5$. Simulations for $\rho=0.4$ are done at $d=1000$. Simulations for $\rho=0.5,0.49$ are done at $d=10000$. Since the dimensionality $d$ is finite in the simulations, effectively $\rho<0.5$ in the numerics. Therefore, simulations always reach a plateau at large $\lambda$.}
\end{figure}

\paragraph*{\textbf{Max-margin and weak regularization} ---} Fig.~\ref{fig:log_vs_hinge} illustrates the generic property that all non-regularized monotone non-increasing loss functions converge to the max-margin solution for linearly separable data \cite{rosset2004margin}. 
%This also holds in our setting, and can be seen explicitly from the analytic formula, see appendix. %Lenka: This is the part we did not manage to show so far. 
Fig.~\ref{fig:log_vs_hinge} depicts a very slow convergence towards this result as a function of regularization parameter $\lambda$ for the logistic loss. While for $\alpha>\alpha^*$ both the hinge and logistic losses performance is basically indistinguishable from the asymptotic one already at $\log \lambda \approx -3$, for $\alpha<\alpha^*$ the convergence of the logistic loss still did not happen even at $\log \lambda \approx -10$. 

\paragraph*{\textbf{Cluster sizes and regularization} ---}
In Fig.~\ref{fig:gen_vs_L_L2} we study in greater detail the dependence of the generalization error both on the regularization $\lambda$ and $\rho$ as $\rho \to 0.5$. We see that the optimality of $\lambda \to \infty$ holds only strictly at $\rho=0.5$ and at any $\rho$ only close to $0.5$ the error at $\lambda \to \infty$ is very large and there is a well delimited region of $\lambda$ for which the error is close to (but strictly above) the Bayes-optimal error. As $\rho \to 0.5$ this interval is getting longer and longer until it diverges at $\rho=0.5$. It needs to be stressed that this result is asymptotic, holding only when $n,d\to \infty$ while $n/d = \alpha$ is fixed. The finite size fluctuations cause that finite size system behaves rather as if $\rho$ was close but not equal to $0.5$, and at finite size if we set $\lambda$ arbitrarily large then we reach a high generalization error. We instead need to optimize the value of $\lambda$ for finite sizes either by cross-validation or otherwise.

\begin{figure}[ht]
\vspace{5mm}
\includegraphics[scale=0.66]{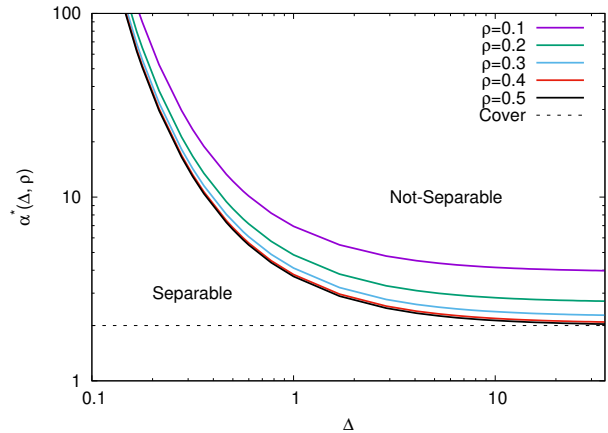}
 \vspace{-3mm}
 \caption{\label{fig:transition} Critical value $\alpha=\alpha^*$, define by Proposition \ref{prop:phase_transition},  at which the linear separability transition occurs as a function of $\Delta$, for different values of $\rho$.
Similarly as for what happens for Gaussian data \cite{candes2018phase}, the MLE does not exists on the left the curve. The line indicates the location of the  transition from linearly separable to non-linearly separable data, that depends on the data structure (the variance $\Delta$ and the fraction $\rho$).}
\end{figure}

\paragraph*{\textbf{Separability phase transition} ---}  The position of the ``interpolation" threshold when data become linearly separable has a well defined limit in the high-dimensional regime as a function of the ratio between the number of samples $n$ and the dimension $d$. The kink in generalization indeed occurs at a value $\alpha^*$ when the training loss of logistic and hinge losses goes to zero (while for the square loss the peak appears at $d=n$ when the system of $n$ linear equations with $d$ parameters becomes solvable). 
The position of $\alpha^*$, given by Proposition \ref{prop:phase_transition}  is shown in  Fig.~\ref{fig:transition} as a function of the cluster variance for different values of $\rho$. For very large cluster variance, the data become random and hence $\alpha= 2$ for equal-sized cluster, as famously derived in classical work by  \cite{cover1965geometrical}. When $\rho<1/2$, however, it is easier to separate linearly the data points and the limiting value of $\alpha^*$ gets larger and differ from Cover's. For finite $\Delta$, the two Gaussian distributions become distinguishable, and the data acquires structure. Consequently, the
$\alpha^*$ is growing as the correlations make data easier to linearly separate again, similarly as described \cite{candes2018phase}.  This phenomenology of the separability phase transition, or equivalently of the existence of the maximum likelihood estimator, thus seems very generic.

% Separability-to-non separability transition

\FloatBarrier

\section*{Acknowledgements}
We thank Pierfrancesco Urbani, Federica Gerace, and Bruno Loureiro for many clarifying discussions related to this project. 
This work is supported by the ERC under the European Union’s Horizon
2020 Research and Innovation Program 714608-SMiLe, by the
French Agence Nationale de la Recherche under grant
ANR-17-CE23-0023-01 PAIL and ANR-19-P3IA-0001 PRAIRIE, and by the US National Science Foundation under grants CCF-1718698 and CCF-1910410. We also
acknowledge support from the chaire CFM-ENS  ``Science des donn\'ees''.
Part of this work was done when Yue Lu was visiting Ecole Normale as a CFM-ENS ``Laplace" invited researcher.
%We gratefully acknowledge the support of NVIDIA Corporation with the
%donation of the Titan Xp GPU used for this research. 
We thank Google Cloud for providing us access to their platform through the Research Credits Application program.

%\newpage
\bibliography{article,refs}
\newpage
\appendix
\def\T{\text{T}}
\def\v{\text{v}}
\def\P{\text{P}}
\def\o{\text{o}}
\def\out{\text{out}}
\def\erfcx{\text{erfcx}}
\def\DD{\text{D}}
\def\spam{\text{spam}}
\def\dd{\text{d}}
\def\argmin{\text{argmin}}
\def\argmax{\text{argmax}}
\def\MSE{\text{MSE}}
\def\MMSE{\text{MMSE}}
\def\iid{\text{i.i.d.}}
\def\eff{\text{eff}}
\def\e{\text{e}}
\def\sign{\text{sign}}
\def\new{\text{new}}
\def\erf{\text{erf}}
\def\gen{\text{gen}}
\def\erfc{\text{erfc}}
\def\bv{{\bf v}}
\def\x{{\bf x}}
\def\w{{\bf w}}
\def\X{{\bf X}}
\def\by{{\bf y}}

%\begin{document}

%\preprint{APS/123-QED}

%\title{The role of regularization
%in classification of high-dimensional noisy Gaussian mixture: supplementary material}

%\maketitle

\onecolumngrid
\numberwithin{equation}{section}

\section{Derivation of the generalization error formula}
\label{app:generalization}
The generalization error is defined as the average fraction of mislabeled instances
\begin{equation}
\varepsilon_{\gen}=\frac{1}{4}\mathbb{E}_{y_{\new},\x_{\new}, \X,\by}\left[\left(y_{\new}-\hat{y}_{\new}\right)^2\right],\label{eq:err_def}
\end{equation}
where $y_{\new}$ is the label of a new observation $\x_{\new}$, and the estimator $\hat{y}_{\new}$ is computed as 
\begin{equation}\hat{y}_{\new}=\sign\left(\frac{\w^\top \x_{\new}}{\sqrt{d}}+b\right).\label{eq:estimator}\end{equation}
Eq. \eref{estimator} holds for every vector $\w=\w\left(\X,\by\right)$ and bias $b=b\left(\X,\by\right)$ computed on the training set $\left\{\X,\by\right\}$. 
Using the fact that $y_{\new},\hat{y}_{\new}= \pm 1$, it is easy to show that \eref{err_def} can be rewritten as
\begin{equation}
\varepsilon_{\gen}=\frac{1}{2}\left(1-\mathbb{E}_{y_{\new},\x_{\new}, \X,\by}\left[y_{\new}\hat{y}_{\new}\right]\right)=\frac{1}{2}\left(1-\mathbb{E}_{y_{\new},\x_{\new},\X,\by}\left[y_{\new}\sign\left(\frac{\w^\top \x_{\new}}{\sqrt{d}}+b\right)\right]\right).\label{eq:egen_1}
\end{equation}
Let us consider the last term in \eref{egen_1}. Using again $y_{\new}=\pm 1$, we can move $y_{\new}$ inside the argument of the sign function and rewrite
\begin{equation}
\mathbb{E}_{y_{\new},\x_{\new},\X,\by}\left[y_{\new}\sign\left(\frac{\w^\top \x_{\new}}{\sqrt{d}}+b\right)\right]=\mathbb{E}_{y_{\new},\x_{\new},\X,\by}\left[\sign\left(\frac{y_{\new}\w^\top \x_{\new}}{\sqrt{d}}+y_{\new}b\right)\right].
\end{equation}
The term $y_{\new}\x_{\new}$ can be rewritten as
\begin{equation}
y_{\new}\x_{\new} =y_{\new}\left(y_{\new}\frac{\bv^\ast}{\sqrt{d}}+\sqrt{\Delta}{\bf z}_{\new}\right)=\frac{\bv^\ast}{\sqrt{d}}+\sqrt{\Delta}{\bf z}'_{\new},
\end{equation}
where ${\bf z}'_{\new}=y_{\new}{\bf z}_{\new} \sim \mathcal{N}({\bf 0},{\bf I}_d)$ has the same distribution as ${\bf z}_{\new}$, since $y_{\new}$ and ${\bf z}_{\new}$ are independent. Hence
\begin{equation}
\mathbb{E}_{y_{\new},\x_{\new},\X,\by}\left[\sign\left(\frac{\w^\top y_{\new}\x_{\new}}{\sqrt{d}}+y_{\new}b\right)\right]\\=\mathbb{E}_{y_{\new},{\bf z}'_{\new},\bv^\ast,\X,\by}\left[\sign\left(\frac{\w^\top \bv^\ast}{d}+\sqrt{\frac{\Delta}{d}}\w^\top{\bf z}'_{\new}+y_{\new}b\right)\right].
\end{equation}
The estimator $\w$ only depends on the training set, hence $\w$ and ${\bf z}'_{\new}$ are independent. We call their rescaled scalar product $\varsigma$, a random variable distributed as a standard normal
\begin{equation}
\varsigma=\frac{1}{\norm{{\bf w}}}{\bf w}^\top {\bf z}'_{\new} \sim \mathcal{N}\left(0,1\right).
\end{equation} 
By averaging over $\varsigma$, we obtain
\begin{equation}
\begin{split}
\mathbb{E}_{y_{\new},{\bf v^\ast},{\bf X},{\bf y},\varsigma}\left[\sign\left(\frac{{\bf w}^\top {\bf v^\ast}}{d}+\sqrt{\frac{\Delta}{d}}\norm{{\bf w}}\varsigma+y_{\new}b\right)\right]\\=\mathbb{E}_{y_{\new},{\bf v^\ast},{\bf X},{\bf y},\varsigma}\left[\sign\left(\frac{1}{\sqrt{\Delta}}\frac{{\bf w}}{||{\bf w}||}^\top \frac{{\bf v^\ast}}{\sqrt{d}}+\varsigma+y_{\new}b\frac{\sqrt{d}}{\sqrt{\Delta}||{\bf w}||}\right)\right],
\end{split}
\end{equation}
where we have used that $\sqrt{\frac{\Delta}{d}}\norm{{\bf w}}>0$ to rescale the argument of the sign function. Finally, we obtain
\begin{equation}
    \begin{split}
\varepsilon_{{\rm gen}}=\frac{1}{2}\left(1-\mathbb{E}_{y_{\new},{\bf v^\ast},{\bf X},{\bf y}}\left[\mathbb{P}\left(\varsigma>-\tau\right)-\mathbb{P}\left(\varsigma<-\tau\right)\right]\right)=\mathbb{E}_{y_{\new},{\bf v^\ast},{\bf X},{\bf y}}\left[Q(\tau)\right].
\end{split}
\end{equation}
where $Q(x) = \frac{1}{\sqrt{2\pi}} \int_x^\infty e^{-t^2/2} dt$ is the Gaussian tail function, and we have defined
\begin{equation}
\tau = \frac{\sqrt{d}}{\sqrt{\Delta}||{\bf w}||}\left(\frac{{\bf w}^\top {\bf v^\ast}}{d}+y_{\new}b\right).
\end{equation}
In the large $d$ limit, the overlaps concentrate to deterministic quantities:
\begin{eqnarray}
\frac{{\bf w}^\top {\bf v^\ast}}{d} \underset{d\rightarrow \infty}{\longrightarrow} m ,\\
\frac{||{\bf w}||}{\sqrt{d}} \underset{d\rightarrow\infty}{\longrightarrow} \sqrt{q}.
\end{eqnarray}
Hence the generalization error reads
\begin{equation}
\varepsilon_{\gen}=\rho Q\Big(\frac{m + b}{ \sqrt{\Delta q}}\Big) + (1-\rho) Q\Big(\frac{m - b}{ \sqrt{\Delta q}}\Big),\label{eq:err_gen}
\end{equation}
where $\rho \in (0,1)$ is the probability that $y_{\new}=+1$.

\section{Derivation of the Bayes-optimal error}
\label{app:BO}
In order to compute the Bayes-optimal error, we consider the posterior distribution of a new label $y_{\new}$, given the corresponding new data point $\bf x_{\new}$ and the estimate $ \bf v$ of the true centroid $\bf v^*$ 
\begin{equation}
{\rm p}\left(y_{\new}|{\bf x}_{\new},{\bf v}\right)\propto {\rm p}\left({\bf x}_{\new} | y_{\new},\bf v\right){\rm p}_y\left(y_{\new}\right)\propto \exp\left(-\frac{1}{2\Delta}\sum_{i=1}^d\left(x_{\new}^i-\frac{y_{\new}\v^i}{\sqrt{d}}\right)^2\right){\rm p}_y(y_{\new}),
\end{equation}
where ``$\propto$" takes into account the normalization over $y_{\new}$.
Similarly, the posterior on ${\bf v}$ given the training data is 
\begin{equation}
{\rm p}\left({\bf v}|{\bf X},{\bf y}\right)\propto {\rm p}\left({\bf X} |\bf v, \bf y\right){\rm p}_{\bf v}\left(\bf v\right) \propto \left[\prod_{\mu=1}^n \exp\left(-\frac{1}{2\Delta}\sum_{i=1}^d\left(x^i_{\mu}-\frac{y_{\mu}\v^i}{\sqrt{d}}\right)^2\right)\right] \exp\left(-\frac{1}{2}\sum_{i=1}^d (\v^i)^2 \right),
\end{equation}
where we remind that ${\bf v}$ has i.i.d. components taken in $\mathcal{N}(0,1)$, and ``$\propto$" takes into account the normalization over ${\bf v}$. We would like to find an explicit expression for
\begin{equation}
{\rm p}\left(y_{\new}|{\bf x}_{\new},{\bf X},{\bf y}\right)= \mathbb{E}_{{\bf v}|{\bf X},{\bf y}}\left[{\rm p}\left(y_{\new}|{\bf x}_{\new},{\bf v}\right)\right],
\end{equation}
in order to estimate the new label as 
\begin{equation}
\hat{y}_{\new} = \arg\,\underset{y' =\pm 1} {\max} \,\log {\rm p}\left(y'|{\bf x}_{\new},{\bf X},{\bf y}\right) .
\end{equation}
Therefore, we have to compute
\begin{equation}
\begin{split}
\mathbb{E}_{{\bf v}|{\bf X},{\bf y}}\left[{\rm p}\left(y_{\new}|{\bf x}_{\new},{\bf v}\right)\right]\propto{\rm p}_y\left(y_{\new}\right)\int \left(\prod _{i=1}^d\dd\v^i\enskip
\e^{-\frac{1}{2} (\v^i )^2} \right) \prod_{\mu=0}^n \e^{-\frac{1}{2\Delta}\sum_{i=1}^d\left(x^i_{\mu}-\frac{y_{\mu}\v^i}{\sqrt{d}}\right)^2},\label{eq:avg_v}
\end{split}
\end{equation}
where in the product over $\mu$ on the right-hand side we have used the notation $y_0=y_{\new}$, ${\bf x}_0={\bf x}_{\new}$. Let us call $I_\v$ the integral over $\bf v$ in \eref{avg_v}.
\begin{equation}
I_\v = \int \prod _{i=1}^d\dd \v^i\enskip
 \e^{-\sum_{i=1}^d\left[\frac{1}{2\Delta}\sum_{\mu=0}^n\left(x^i_{\mu}-\frac{y_{\mu}\v^i}{\sqrt{d}}\right)^2+\frac{1}{2} (\v^i )^2\right]}=\prod_{i=1}^d \int \dd \v\enskip
 \e^{-\frac{1}{2\Delta}\sum_{\mu=0}^n\left(x^i_{\mu}-\frac{y_{\mu}\v}{\sqrt{d}}\right)^2-\frac{1}{2} \v^2},
\end{equation}
where in the last equality we have dropped the index $i$ from the components of $\bf v$ for simplicity, since they are all independent.
Computing the integral over $\v$, we obtain 
\begin{equation}
\begin{split}
I_\v = C\left(\alpha,\Delta,d\right) \prod _{i=1}^d\prod_{\mu=0}^n\exp\left(-\frac{1}{2\Delta \left(\alpha+\Delta+\frac{1}{d}\right)} \left((\alpha+\Delta)(x_{\mu}^i )^2-\frac{\alpha}{n}y_{\mu}x_{\mu}^i \sum_{\underset{\nu\neq\mu}{\nu=0}}^n y_{\nu}x_{\nu}^i \right)\right)\\
= C\left(\alpha,\Delta,d\right) \exp\left(-\frac{1}{2\Delta \left(\alpha+\Delta+\frac{1}{d}\right)}\sum_{i=1}^d \left((\alpha+\Delta)(x_{\new}^i)^2-\frac{\alpha}{n} y_{\new}x_{\new}^i\sum_{\nu=1}^n y_{\nu}x_{\nu}^i \right)\right)\\
\times \exp\left(-\frac{1}{2\Delta \left(\alpha+\Delta+\frac{1}{d}\right)}\sum_{\mu=1}^n\sum_{i=1}^d \left((\alpha+\Delta)(x_{\mu}^i)^2-\frac{\alpha}{n} y_{\mu}x_{\mu}^i\sum_{\underset{\nu\neq\mu}{\nu=1}}^n y_{\nu}x_{\nu}^i -\frac{\alpha}{n}y_{\mu}x_{\mu}^i y_{\new}x_{\new}^i\right)\right)\\
=  C\left(\alpha,\Delta,d\right) \tilde{C}\left({\bf X}, {\bf y},{\bf x}_{\new},\alpha,\Delta,d\right) \exp\left(\frac{\alpha}{\Delta \left(\alpha+\Delta+\frac{1}{d}\right)}  y_{\new}{\bf x}_{\new} ^\top \frac{1}{n}\sum_{\mu=1}^n y_{\mu}{\bf x}_{\mu}  \right),
\end{split}
\end{equation}
where the first two factors $C$ and $\tilde{C}$ contain all the terms that do not depend on $y_{\new}$. Therefore
\begin{equation}
\hat{y}_{\new}=\arg\,\underset{y = \pm 1}{\max}\, \left[\frac{\alpha}{\Delta \left(\alpha+\Delta+\frac{1}{d}\right)}  y{\bf x}_{\new} ^\top\frac{1}{n}\sum_{\mu=1}^n  y_{\mu}{\bf x}_{\mu} +\log {\rm p}_y\left(y\right)\right].\label{eq:yhatBO_supmat}
\end{equation} 
Using the fact that $y_{\mu}{\bf x}_{\mu}=\frac{{\bf v}^*}{\sqrt{d}}+\sqrt{\Delta}{\bf z}_{\mu}$, ${\bf z}_{\mu}\sim \mathcal{N}(0, {\bf I}_d)$ and ${\bf v}^*$ is the true realization of $\bf v$, the first term in \eref{yhatBO_supmat} in the limit where $n,d\rightarrow\infty$ can be rewritten as
\begin{equation}
\begin{split}
\frac{1}{n}\sum_{\mu=1}^n {\bf x}_{\new} ^\top y_{\mu}{\bf x}_{\mu} \underset{n,d \rightarrow\infty}{\longrightarrow}y_{\new}+\sqrt{\Delta\left(1+\frac{\Delta}{\alpha}\right)} z'_{\new},\label{eq:avg_overlap}
\end{split}
\end{equation}
where $z'_{\new}\sim \mathcal{N}(0,1)$. Therefore, in the large $d$ limit we find that
\begin{equation}
\hat{y}_{\new}=\arg\,\underset{y=\pm 1}{\max}\, \left[\frac{\alpha}{\Delta \left(\alpha+\Delta\right)}  y\left(y_{\new}+\sqrt{\Delta\left(1+\frac{\Delta}{\alpha}\right)}  z'_{\new}\right) +\log {\rm p}_y\left(y\right) \right].\label{eq:y_argmax}
\end{equation}
It is useful to rewrite the generalization error as
\begin{equation}
\varepsilon_{\gen} = \frac{1}{4}\mathbb{E}_{\mat{X}, {\bf y}, {\bf x}_{\new}, y_{\new}}\left[(\hat{y}_{\new}-y_{\new})^2\right] = \sum_{y_{\new}=-1,1} \mathbb{P} \left(\hat{y}_{\new}\neq y_{\new}\right){\rm p}_y(y_{\new}).
\end{equation}
Using \eref{y_argmax}, we can compute 
\begin{equation}
\begin{split}
 \mathbb{P} \left(\hat{y}_{\new}\neq y_{\new}\right)
 =\mathbb{P}\left(y_{\new}z'_{\new}<-\sqrt{\frac{\alpha}{\Delta (\alpha+\Delta)}}\left(1+ \left(1+\frac{\Delta}{\alpha}\right)\frac{\Delta}{2}\log\frac{{\rm p}_y(y_{\new})}{{\rm p}_y(-y_{\new})}\right)\right).\label{eq:prob_BO_err}\end{split}
 \end{equation}
 If $y_{\new}=1$, \eref{prob_BO_err} gives
 \begin{equation}
  \mathbb{P} \left(\hat{y}_{\new}\neq 1\right)=Q\left(\frac{\frac{\alpha}{\Delta+\alpha}+\frac{\Delta}{2}\log\frac{\rho}{1-\rho}}{\sqrt{\Delta \frac{\alpha}{\Delta+\alpha}}}\right),
 \end{equation}
 where $Q(x) = \frac{1}{\sqrt{2\pi}} \int_x^\infty e^{-t^2/2} dt$ is the Gaussian tail function.
  If $y_{\new}=-1$, \eref{prob_BO_err} gives
 \begin{equation}
  P \left(\hat{y}_{\new}\neq -1\right)=Q\left(\frac{\frac{\alpha}{\Delta+\alpha}-\frac{\Delta}{2}\log\frac{\rho}{1-\rho}}{\sqrt{\Delta \frac{\alpha}{\Delta+\alpha}}}\right).
 \end{equation}
Using the fact that $\rho={\rm p}_y(1)$ and $1-\rho={\rm p}_y(-1)$, we get that
 \begin{equation}
 \varepsilon_{\gen}^{\rm BO}=\rho Q\left(\frac{\frac{\alpha}{\Delta+\alpha}+\frac{\Delta}{2}\log\frac{\rho}{1-\rho}}{\sqrt{\Delta \frac{\alpha}{\Delta+\alpha}}}\right)+(1-\rho)Q\left(\frac{\frac{\alpha}{\Delta+\alpha}-\frac{\Delta}{2}\log\frac{\rho}{1-\rho}}{\sqrt{\Delta \frac{\alpha}{\Delta+\alpha}}}\right).\label{eq:BOerror}
 \end{equation}
 \subsection{Bayes-optimal estimator}
It is worth noting that the optimal error in \eref{BOerror} can be achieved by the plug-in estimator
\begin{equation}
    {\bf \hat w}=\frac{\sqrt{d}}{n}\sum_{\mu=1}^n y_{\mu}{\bf x}_{\mu}\label{eq:plugin_w_supmat}.
\end{equation}
This result was already shown in \cite{lelarge2019asymptotic} for the case of symmetric clusters. The optimal bias is obtained from the minimization of the generalization error \eref{err_gen} with respect to $b$, at fixed $m,q$. This yields:
\begin{equation}
\hat b =\underset{b}{\argmin}\enskip\varepsilon_{\rm gen}(q,m)=\frac{q}{m} \frac{\Delta} {2}\log \left(\frac{\rho}{1-\rho}\right).
\end{equation}
Substituting \eref{plugin_w_supmat} in the definition of the overlaps \eref{overlaps} in the main text, we obtain that the values of $m$ and $q$ associated to the plugin estimator are 
\begin{equation}
    m=1 ,\qquad q=\left(1+\tfrac{\Delta}{\alpha}\right).
\end{equation}
Hence, the generalization error of the plug-in estimator is
\begin{equation}
\begin{split}
    \varepsilon_{\rm gen}^{\rm plugin}=\mathbb{P} \left(y_{\new}\left(\frac{1}{\sqrt{d}}{\bf \hat w}^\top {\bf x}_{\new}+\hat b\right)<0\right)\\
    = \mathbb{P}\left(y_{\new} z'_{\new}<-\sqrt{\frac{\alpha}{\Delta (\alpha +\Delta)}}\left(1+y_{\new} \left(1+\frac{\Delta}{\alpha}\right)\frac{\Delta}{2}\log\frac{\rho}{1-\rho}\right)\right),\label{eq:gen_err_plugin}
\end{split}
\end{equation}
where we have used \eref{avg_overlap} in the last equality. The probability in \eref{gen_err_plugin} is the same as in \eref{prob_BO_err}. Hence, the plug-in estimator achieves the Bayes-optimal error.

\section{Details of proofs}
\label{app:train}

In what follows, we provide more technical details for several key results stated in \sref{sketches}. They serve as the basis of the proof of Proposition~\ref{prop:Gordon_det}.

\subsection{Proof of Proposition~\ref{prop:Gordon_E}}
\label{app:proof_Gordon_E}
Recall from the main text that
\[
\begin{aligned}
{\cal L}_\lambda(q, m, b) &=\frac{\lambda q}{2} + \underset{{\bf w} \in {\cal S}_{q, m}}{\min}\!\! \max_{{\bf u}} \frac{1}{d}\sum_{i=1}^n \Big[u_i \! \Big(\frac{{\bf w}^\top {\bf v}^\ast}{d} \!\! +\!\!  \sqrt{\Delta} \frac{y_i {\bf z}_i^\top {\bf w}}{\sqrt{d}} + b y_i\Big) \!\! - \widehat{\ell}(u_i)\Big]\\
&=\frac{\lambda q}{2} + \underset{{\bf w} \in {\cal S}_{q, m}}{\min}\!\! \max_{{\bf u}} \frac{1}{d}\sum_{i=1}^n \Big[u_i (m+b y_i) - \widehat{\ell}(u_i) + \sqrt{\frac{\Delta}{d}} u_i y_i {\bf z}_i^\top {\bf w}\Big],
\end{aligned}
\]
where in reaching the second equality we have used the fact that any ${\bf w} \in {\cal S}_{q, m}$ satisfies the equality $m = \frac{1}{d} {\bf w}^\top {\bf v}^\ast$. Introduce an auxiliary problem
\[
\begin{aligned}
\widetilde{\cal L}_\lambda(q, m, b) &= \frac{\lambda q}{2} + \underset{{\bf w} \in {\cal S}_{q, m}}{\min}\!\! \max_{{\bf u}}\set{\frac{1}{d}\sum_{i=1}^n \Big[u_i (m+b y_i) - \widehat{\ell}(u_i)\Big] + \sqrt{\frac{\Delta}{d}} \norm{{\bf u}} \frac{{\bf g}^\top {\bf w}}{d} + \sqrt{\Delta q} \Big(\frac{1}{d}\sum_{i=1}^n u_i y_i s_i\Big)}\\
&= \frac{\lambda q}{2} + \underset{{\bf w} \in {\cal S}_{q, m}}{\min}\!\! \max_{{\bf u}} \set{\frac{1}{d}\sum_{i=1}^n \Big[u_i h_i - \widehat{\ell}(u_i)\Big] + \sqrt{\frac{\Delta}{d}} \norm{{\bf u}} \frac{{\bf g}^\top {\bf w}}{d}},
\end{aligned}
\]
where ${\bf g} = (g_1, g_2, \ldots, g_d)^\top$ and ${\bf s} = (s_1, s_2, \ldots, s_n)$ are two independent random vectors whose entries are drawn from the i.i.d. standard normal distribution, and $h_i = \sqrt{\Delta q} (y_i s_i) + m + b y_i$. As $y_i \in \set{\pm 1}$, independent of $s_i$, we note that $h_i$ has the same probability distribution as the quantity defined in \eref{h_vec} in the main text.

Gordon's minimax inequalities \cite{Gordon:85,Gordon:1988lr,chris:152} allow us to make the following comparison: For any constants $c$ and $\delta > 0$, we have
\begin{equation}\label{eq:Gordon_minmax}
\mathbb{P}({\cal L}_\lambda(q, m, b) < c) \le 2 \, \mathbb{P}(\widetilde{\cal L}_\lambda(q, m, b) < c).
\end{equation}
To connect this to the statements in Proposition~\ref{prop:Gordon_E}, we note that
\[
\begin{aligned}
\widetilde{\cal L}_\lambda(q, m, b) &\ge \frac{\lambda q}{2} + \max_{{\bf u}} \!\!\underset{{\bf w} \in {\cal S}_{q, m}}{\min}  \set{\frac{1}{d}\sum_{i=1}^n \Big[u_i h_i - \widehat{\ell}(u_i)\Big] + \sqrt{\frac{\Delta}{d}} \norm{{\bf u}} \frac{{\bf g}^\top {\bf w}}{d}}\\
&=  \frac{\lambda q}{2} + \max_{{\bf u}}   \set{\frac{1}{d}\sum_{i=1}^n \Big[u_i h_i - \widehat{\ell}(u_i)\Big] - \sqrt{\frac{\Delta \norm{{\bf u}}^2 (q-m^2)}{d}} \frac{\norm{{\bf g}}}{\sqrt{d}}}\\
&= {\cal E}^{(d)}_\lambda(q, m, b).
\end{aligned}
\]
It follows that
\[
\mathbb{P}(\widetilde{\cal L}_\lambda(q, m, b) < c) \le \mathbb{P}({\cal E}^{(d)}_\lambda(q, m, b) < c).
\]
Combining this inequality with \eref{Gordon_minmax} gives us the first inequality in Proposition~\ref{prop:Gordon_E}. To obtain the second inequality in the proposition, we use the fact that the unconstrained optimization problem in \eref{tl} for the global training loss ${\cal L}^\ast$ is convex. Following exactly the same strategy as used in \cite{chris:152}, we can interchange the order of $\min$ and $\max$ in the dual formulation of \eref{tl}, which then allows us to reach the two-sided inequality in \eref{Gordon_E2}. 

\subsection{Proof of Lemma~\ref{lemma:Lagrange}}

We first rewrite the optimization problem in \eref{E_vec} as
\begin{equation}\label{eq:E_vec_opt}
\max_{\mu \ge 0} \ \max_{\norm{\vu}^2/d = \mu} \set{-\sqrt{\Delta_d\mu(q - m^2)} + \frac{\vu^\top \vh}{d} -\frac{1}{d} \sum_{i=1}^n \widetilde{\ell}(u_i)}.
\end{equation}
For the inner maximization, the constraint on the squared norm $\norm{{\bf u}}^2$ weakly couples different coordinates of $\vu$ together. To fully decouple these coordinates, we introduce a Lagrangian function
\[
\frac{\vu^\top \vh}{d} -\frac{1}{d} \sum_{i=1}^n \widetilde{\ell}(u_i) - \frac{\gamma}{2d}(\norm{{\bf u}}^2 - \mu d),
\]
where $\gamma > 0$ is the Lagrange multiplier. For any fixed $\gamma$, the optimal solution $\vu_\gamma \in \R^n$ can be obtained by setting the gradient of the Lagrangian function to zero, which gives us
\[
\nabla \widetilde{\ell}(\vu_\gamma) + \gamma \vu_\gamma = \vh.
\]
Since there is a one-to-one correspondence between the Lagrange multiplier $\gamma$ and the normalized squared norm $\mu = \norm{\vu_\gamma}^2/ d$, it is thus equivalent to solve \eref{E_vec_opt} in terms of 
\[
\max_{\gamma > 0}\Big\{-\sqrt{\frac{\Delta_d(q-m^2) \norm{\vu_\gamma}^2}{d}} + \frac{\vu_\gamma^\top h}{d} - \frac{1}{d} \sum_{i=1}^n \widetilde{\ell}(u_{\gamma,i})\Big\}
\]
and thus we get \eref{E_vec_d}.

\subsection{Proof of Proposition~\ref{prop:Gordon_det}}

We first establish \eref{comp_phi_e} for the special case where the subset $\Omega$ is a singleton. In this case, we just need to show
\begin{equation}\label{eq:conv_point}
\mathbb{P}\Big({\cal L}_\lambda(q, m, b) \ge {\cal E}_\lambda(q, m, b) - \delta\Big) \to 1.
\end{equation}
for any fixed $q, m$ and $b$. 
\\
Recall the characterization of ${\cal E}^{(d)}_\lambda(q, m, b)$ given in Lemma~\ref{lemma:Lagrange}. The problem in \eref{E_vec_d} reaches its maximum at a point $\gamma^\ast_d$ where the derivative of the function to be maximized is equal to 0. In calculating this derivative, we need the quantity $\frac{d u_{\gamma,i}}{d \gamma}$, which can be obtained as
\[
\widetilde{\ell}''(u_{\gamma, i}) \frac{d u_{\gamma,i}}{d \gamma} + u_{\gamma,i} + \gamma \frac{d u_{\gamma,i}}{d \gamma}  = 0
\]
and thus $\frac{d u_{\gamma,i}}{d \gamma}  = \frac{-u_{\gamma, i}}{\widetilde{\ell}''(u_{\gamma, i}) + \gamma}$. Using this expression and after some simple manipulations, we get 
\begin{equation}\label{eq:fix_point_d_sup}
\alpha (\gamma^\ast_d)^2 \frac{\norm{\vu_{\gamma^\ast_d}}^2}{n} = \Delta_d (q- m^2).
\end{equation}
Moreover,
\begin{equation}\label{eq:E_s_d_sup}
{\cal E}^{(d)}_\lambda(q, m, b) = \frac{\sum_{i=1}^n \big[u_{\gamma^\ast_d, i} \widetilde{\ell}'(u_{\gamma^\ast_d,i}) - \widetilde{\ell}(u_{\gamma^\ast_d,i})\big]}{d} + \frac{\lambda q}{2}.
\end{equation}

Next, we introduce the following scalar change of variables: $v_{\gamma, i} = \widetilde{\ell}'(u_{\gamma, i})$. It is easy to verify from properties of Legendre transformations that
\[
u_{\gamma, i} = \ell'(v_{\gamma, i}) \quad\text{and}\quad u_{\gamma, i} \widetilde{\ell}'(u_{\gamma, i}) - \widetilde{\ell}(u_{\gamma, i}) = \ell(v_{\gamma, i}).
\]
Substituting these identities, we can characterize $v_{\gamma, i}$ via the implicit equation
\begin{equation}\label{eq:v_vec_h}
v_{\gamma, i} + \gamma \ell'(v_{\gamma, i}) = h_i.
\end{equation}
Moreover, \eref{fix_point_d_sup} can now be rewritten as
\begin{equation}\label{eq:fix_point_gamma_sup}
\alpha (\gamma^\ast_d)^2 \frac{1}{n} \sum_{i=1}^n [\ell'(v_{\gamma^\ast_d, i})]^2 = \Delta_d (q- m^2)
\end{equation}
and more importantly, \eref{E_s_d_sup} can be simplified as
\[
{\cal E}^{(d)}_\lambda(q, m, b) = \frac{\alpha}{n} \sum_{i=1}^n \ell(v_{\gamma^\ast_d, i}) + \frac{\lambda q}{2}.
\]

Let $v_\gamma$ be a random variable defined via the implicit equation 
\begin{equation}\label{eq:v_implicit}
v_\gamma + \gamma \ell'(v_\gamma) = h,
\end{equation}
 where $h = \sqrt{\Delta q} s + m + b y$ with $S \sim \mathcal{N}(0, 1)$ and $y$ being a random variable independent of $s$ such that
\[
\mathbb{P}(y = 1) = \rho \quad \text{and} \quad \mathbb{P}(y = -1) = 1-\rho.
\]
Since the loss function $\ell(\cdot)$ is convex, the function $v + \gamma \ell'(v)$ is strictly increasing. It follows that the distribution function of $v_\gamma$ is given as in \eref{v_CDF}. As $n, d \to \infty$ with $d/n$ fixed at $\alpha$, we have $\Delta_d \to \Delta$ and 
\[
\frac{1}{n} \sum_{i=1}^n [\ell'(v_{\gamma, i})]^2 \to \mathbb{E}[(\ell'(v_\gamma))^2]
\]
\emph{uniformly} over any compact subset of $\gamma$. It follows that $\gamma^\ast_d$ as defined in \eref{fix_point_gamma_sup} converges to $\gamma^\ast$, which is the unique solution of \eref{fix_point_gamma}. Moreover, we have
\begin{equation}\label{eq:Ed_E_det}
{\cal E}^{(d)}_\lambda(q, m, b) \to {\cal E}_\lambda(q, m, b) = \alpha \mathbb{E}[\ell(v_{\gamma^\ast})] + \frac{\lambda q}{2}.
\end{equation}

For any $\delta > 0$, we can apply Proposition~\ref{prop:Gordon_E} to get
\[
\mathbb{P}({\cal L}_\lambda(q, m, b) < {\cal E}_\lambda(q, m, b) - \delta) \le 2 \mathbb{P}({\cal E}^{(d)}_\lambda(q, m, b) < {\cal E}_\lambda(q, m, b) - \delta).
\]
As the right-hand side tends to $0$ due to \eref{Ed_E_det}, we have \eref{conv_point}.

Let $\Omega$ be an arbitrary compact subset of $\set{(q, m, b): m^2 \le q}$. We denote by $\Omega_K$ a finite subset of $\Omega$ consisting of $K$ points, \emph{i.e.}, $\Omega_K = \set{(q_k, m_k, b_k) \in \Omega: 1 \le k \le K}$.
\[
\begin{aligned}
\mathbb{P}({\cal L}_\lambda(\Omega_K) < {\cal E}_\lambda(\Omega) - \delta) &= \mathbb{P}(\cup_{k=1}^K \set{{\cal L}_\lambda(q_k, m_k, b_k) < {\cal E}_\lambda(\Omega) - \delta})\\
&\le \sum_{k=1}^K \mathbb{P}({\cal L}_\lambda(q_k, m_k, b_k) < {\cal E}_\lambda(\Omega) - \delta)\\
&\le \sum_{k=1}^K \mathbb{P}({\cal L}_\lambda(q_k, m_k, b_k) < {\cal E}_\lambda(q_k, m_k, b_k) - \delta).\\
\end{aligned}
\]
As $n \to \infty$, the right-hand side of the inequality tends to $0$. It follows that $\mathbb{P}({\cal L}_\lambda(\Omega_K) \ge {\cal E}_\lambda(\Omega) - \delta) \to 1$. Note that this characterization holds for any finite $K$. From the smoothness of the optimization problem \eref{tl_mq}, one can construct a family of subsets $\set{\Omega_K}$ such that ${\cal L}_\lambda(\Omega_K) \to {\cal L}_\lambda(\Omega)$ as $K \to \infty$, and thus we have \eref{comp_phi_e}. This strategy follows closely the approach used in \cite{chris:152}. Finally, to get \eref{comp_phi_e_ast}, we first note that \eref{comp_phi_e} implies that
\[
\mathbb{P}\Big({\cal L}_\lambda^\ast \ge {\cal E}_\lambda^\ast - \delta\Big) \to 1.
\]
The ``other direction'', \emph{i.e.}, $\mathbb{P}\Big({\cal L}_\lambda^\ast \le {\cal E}_\lambda^\ast + \delta\Big) \to 1$ can be obtained by exploiting the convexity of the loss function $\ell(\cdot)$, which allows us to interchange the order of $\min$ and $\max$ in the dual formulation of \eref{tl}. We omit the details as they follow exactly the same strategy as used in \cite{chris:152}. 

\subsection{Proof of Proposition~\ref{prop:phase_transition}}
\label{app:proof_phase_transition}
We start with the fixed-point equation for the Lagrange multiplier given in \eref{fix_point_gamma}. For our proof, it will be more convenient to rewrite this equation in terms of the random variable $u_\gamma \bydef \ell'(v_\gamma)$. It is a well-known property of Legendre transformations that we can write the ``symmetric equation'' $v_\gamma = \widetilde{\ell}'(u_\gamma)$. Since $v_\gamma$ is determined via the implicit equation \eref{v_implicit}, we have
\[
\widetilde{\ell}'(u_\gamma) + \gamma u_\gamma = h.
\]
It follows that the cumulant distribution function of $u_\gamma$ is given by
\[
\mathbb{P}(u_\gamma \le u) = \rho Q\left(\frac{\widetilde{\ell}'(u) + \gamma u - m - b}{\sqrt{\Delta q}}\right)+ (1-\rho)Q\left(\frac{\widetilde{\ell}'(u) + \gamma u - m + b}{\sqrt{\Delta q}}\right),
\]
where $Q(\cdot)$ is the distribution function of a standard normal random variable. Writing \eref{fix_point_gamma} in terms of $u_\gamma$, we have
\begin{equation}\label{eq:fix_point_u_sup}
\alpha \gamma^2 \mathbb{E}[ u_\gamma^2] = \Delta (q- m^2).
\end{equation}

Our assumption of the loss function $\ell(\cdot)$ is that it is convex and monotonically decreasing, with $\ell(+\infty) = \ell'(+\infty) = 0$. It follows that $\ell'(-\infty) < u_\gamma < 0$. Introducing the changes of variables $\theta \bydef m / \sqrt{q}$, $\widetilde{b} \bydef b / \sqrt{q}$ and $\widetilde{\gamma} = \gamma/\sqrt{q}$, and using the identity $\mathbb{E}[ u_\gamma^2] =(-2) \int_{\ell'(-\infty)}^0 u \mathbb{P}(u_\gamma \le u) du$, we can rewrite \eref{fix_point_u_sup} as
\begin{equation}\label{eq:fix_point_u_sup2}
\alpha S(\widetilde{\gamma}, q, \theta) = \Delta(1-\theta^2),
\end{equation}
where
\[
S(\widetilde{\gamma}, q, \theta) \bydef \widetilde{\gamma}^2 \int_0^{-\ell'(-\infty)} (2u) \left(\rho Q\Big(\frac{\widetilde{\ell}'(-u)}{\sqrt{\Delta q}} +\frac{ -\widetilde{\gamma} u - \theta - \widetilde{b}}{\sqrt{\Delta}}\Big)+ (1-\rho)Q\Big(\frac{\widetilde{\ell}'(-u)}{\sqrt{\Delta q}} +\frac{ -\widetilde{\gamma} u - \theta + \widetilde{b}}{\sqrt{\Delta}}\Big)\right) du.
\]
We further denote by $\widehat{\gamma}^\ast(q, \theta)$ the solution to \eref{fix_point_u_sup2}. We can show that, for any fixed $\widetilde{\gamma}$ and $\theta$, the function $S(\widetilde{\gamma}, q, \theta)$ is monotonically decreasing as we increase $q$. Moreover,
\[
\lim_{q \to \infty} S(\widetilde{\gamma}, q, \theta) = S^\ast(\widetilde{\gamma}, \theta) \bydef \int_0^{-\widetilde{\gamma}\ell'(-\infty)} (2u) \Big[\rho Q\Big(\frac{ -u - \theta - \widetilde{b}}{\sqrt{\Delta}}\Big)+ (1-\rho)Q\Big(\frac{ -u - \theta + \widetilde{b}}{\sqrt{\Delta}}\Big)\Big] du.
\]

Clearly, $S^\ast(\widetilde{\gamma}, \theta)$ is monotonic with respect to $\widetilde{\gamma}$, but it has a finite limit as $\widetilde{\gamma} \to \infty$, \emph{i.e.},
\[
\lim_{\widetilde{\gamma} \to \infty} S^\ast(\widetilde{\gamma}, \theta) = \Delta \int_0^\infty u^2 \Big[\rho f\Big(u+\frac{\theta + \widetilde{b}}{\sqrt{\Delta}}\Big)+ (1-\rho)f\Big(u+\frac{ \theta - \widetilde{b}}{\sqrt{\Delta}}\Big)\Big],
\]
where $f(\cdot)$ is the probability density function of $\mathcal{N}(0, 1)$. An implication of this limit being finite is that, although the Lagrange multiplier $\widehat{\gamma}^\ast(q, \theta)$ remains finite for any fixed $q$, it tends to $\infty$ as $q \to \infty$ if 
\begin{equation}\label{eq:alpha_c_sup}
\alpha < \frac{1-\theta^2}{S^\ast(\infty, \theta)}.
\end{equation}
It follows from \eref{v_implicit} that, as $\gamma \to \infty$, $\ell'(v_\gamma) \to 0$ and thus $v_\gamma \to \infty$. Consequently, 
\[
\lim_{q \to \infty} {\cal E}_{\lambda=0}(q, m, b) = \lim_{q \to \infty} \alpha \mathbb{E}[\ell(v_{\gamma^\ast(q, \theta)})] \to 0.
\]
This characterization can be interpreted as follows: If there exists a $\theta$ that satisfies \eref{alpha_c_sup}, then as we move along the ``ray'' of constant slope $\theta = m / \sqrt{q}$, the training loss ${\cal E}_{\lambda=0}(q, m, b)$ will tend to $0$. The critical threshold $\alpha^\ast$ can then be obtained by maximizing the right-hand side of \eref{alpha_c_sup}, which gives us the final expression as stated in Proposition~\ref{prop:phase_transition}.

\subsection{Derivation of Theorem \ref{th1} from Gordon's characterization}
\label{app:proof_mapping}
In this section, we show that the fixed point equations in Theorem \ref{th1} can be mapped to Gordon's characterization, namely \eref{fix_point_gamma} and \eref{comp_phi_e_ast} in the main text. First of all, we observe that \eref{fix_point_gamma} is trivially satisfied by the solution of system \eref{1}-\eref{6}. Then, we consider the minimization of ${\cal E}_\lambda(q, m, b)$, derived in \eref{Ed_E_det}, with respect to $q,m,b$. This simply amounts to setting the derivatives to zero. Note that the partial derivatives of $v$ and $\gamma^*$ can be computed by taking the derivatives of both sides of \eref{v_vec_h} and \eref{fix_point_gamma} respectively. The minimization leads to the following system of equations:
\begin{eqnarray}
\alpha \sqrt{\frac{\Delta}{q}}\mathbb{E}_{y,s}\left[\ell ' (v_{\gamma^{\ast}})s\right]+\lambda=\frac{\Delta}{\gamma},\label{eq:mapping_q}\\
m=-\alpha \frac{\gamma}{\Delta}\mathbb{E}_{y,s}\left[\ell ' (v_{\gamma^{\ast}})\right],\label{eq:mapping_m}\\
\mathbb{E}_{y,s}\left[y\ell ' (v_{\gamma^{\ast}})\right]=0,\label{eq:mapping_b}
\end{eqnarray}
where $s\sim\mathcal{N}(0,1)$, $y=+1$ with probability $\rho\in(0,1)$ and $y=-1$ otherwise.
We observe that \eref{mapping_b} is the same as \eref{11} and \eref{mapping_m} is equivalent to \eref{1} and \eref{gamma}.
Using again \eref{gamma}, we can rewrite \eref{mapping_q} as
\begin{equation}
\hat\gamma=\alpha \sqrt{\frac{\Delta}{q}}\mathbb{E}_{y,s}\left[\ell ' (v_{\gamma^{\ast}}) s\right].
\end{equation}
Note that $\ell ' (v_{\gamma^{\ast}}(h(s)))$ is a function of $s$, and $\ell ''$ is well defined. Therefore, we can apply Stein's lemma and rewrite 
\begin{equation}
\hat\gamma=\alpha \sqrt{\frac{\Delta}{q}} \mathbb{E}_{y,s}\left[\partial_s v_{\gamma^{\ast}} \ell '' (v_{\gamma^{\ast}})\right],
\end{equation}
which leads to an identity if we substitute the definition of $\hat \gamma$ provided in \eref{6}.
\section{Evaluation of the fixed point equations}
\label{app:fixed_point}
In this section we will compute the fixed-point equations for the square and hinge loss. The equations for the logistic loss cannot be computed analytically and require numerical integration.
\subsection{Square loss}
In this case, $\ell (\omega)=\frac{1}{2}(\omega-1)^2$ and the fixed point equations \eref{1}-\eref{6} can be inverted analytically. The minimizer $v$, defined as 
\begin{equation}
      v \equiv \,\underset{\omega}{{\argmin}} \frac{(\omega-h(y,m,q,b))^2}{2\gamma} + \frac{1}{2}(\omega-1)^2,
\end{equation}
is simply
\begin{equation}
    v=h-\gamma l'(v) = \frac{h+\gamma}{1+\gamma},
\end{equation}
where $h\sim \mathcal{N}(m+yb, \Delta q)$. Hence, we obtain
\begin{eqnarray}
\hat m &=& \frac{\alpha}{\gamma} {\mathbb E}_{y,h}\left[v(y,h,\gamma)-h\right]=\frac{\alpha}{1+\gamma}\left(1-m-(2\rho-1)b\right),\label{eq:l2_mhat}\\
\hat q &=& \frac{\alpha \Delta}{\gamma^2} {\mathbb E}_{y,h}\left[(v(y,h,\gamma)-h)^2\right]=\frac{\alpha\Delta}{(1+\gamma)^2}\left(\Delta q + \mathbb{E}_y\left[(1-m-yb)^2\right]\right),\label{eq:l2_qhat}\\
\hat{\gamma} &=& \frac{\alpha \Delta}{\gamma} \left(1 - {\mathbb E}_{y,h}\left[ \partial_{h} v(y,h,\gamma)\right]\right)=\frac{\alpha\Delta}{1+\gamma}.
\label{eq:l2_gammahat}
\end{eqnarray}
To compute the bias $b$, we have to solve
\begin{equation}
    0=\mathbb{E}_{y,h}\left[y(v-h)\right]=\frac{\gamma}{1+\gamma}\mathbb{E}_{y,h}\left[y(1-h)\right],
\end{equation}
which simply gives
\begin{equation}
b=(2\rho-1)(1-m).
\end{equation}
We can plug \eref{l2_mhat}-\eref{l2_gammahat} in the equations for $m,q,\gamma$ to obtain
\begin{eqnarray}
\gamma &=& \frac {\Delta}{\lambda+\hat{\gamma}}=\frac{\Delta(1-\alpha)-\lambda +\sqrt{(\Delta(1-\alpha)-\lambda)^2+4\lambda\Delta}}{2\lambda},\label{eq:l2_gamma}\\
m &=& \frac {\hat m}{\lambda+\hat{\gamma}}=\frac{4\alpha\gamma \rho(1-\rho)}{\Delta(1+\gamma) +4\alpha \gamma \rho(1-\rho)},\label{eq:l2_m}\\
q &=& \frac {\hat q + \hat m^2}{(\lambda+\hat{\gamma})^2}=\frac{1}{(1+\gamma)^2-\alpha \gamma^2}\left(\frac{\alpha \gamma^2}{\Delta}((1-m)^2-b^2)+(1+\gamma)^2m^2\right).\label{eq:l2_q}
\end{eqnarray}

\subsection{Hinge loss}
In this case, $\ell(\omega)=\max\{0,1-\omega\}$ and the minimizer
\begin{equation}
      v \equiv \,\underset{\omega}{{\argmin}} \frac{(\omega-h(y,m,q,b))^2}{2\gamma} + \max\{0,1-\omega\},
\end{equation}
is piece-wise defined as
\begin{equation}
v=
\begin{cases}
h & \text{if}\quad h>1\\
1 & \text{if}\quad 1-\gamma<h<1\\
h+\gamma & \text{if}\quad h<1-\gamma
\end{cases}.
\end{equation}
From \eref{1}-\eref{6}, it follows that
\begin{eqnarray}
 \gamma = \frac{\gamma}{K_{\gamma}},\\
 m=\frac{\alpha}{\Delta}\frac{K_m}{K_{\gamma}},\\
 q=\frac{\alpha}{\Delta K_{\gamma}^2}\left(K_q+\frac{\alpha}{\Delta}K_m^2\right),
\end{eqnarray}
where we have defined
\begin{equation}
K_\gamma = \frac{\lambda \gamma}{\Delta}+\alpha\left(1-\mathbb{E}_y\left[Q\left(\frac{1-m-yb}{\sqrt{\Delta q}}\right)+Q\left(\frac{\gamma-(1-m-yb)}{\sqrt{\Delta q}}\right)\right]\right),
\end{equation}
\begin{equation}
\begin{split}
K_m = \sqrt{\frac{\Delta q}{2\pi}}\mathbb{E}_y \left[\exp\left(-\frac{(1-m-yb)^2}{2\Delta q}\right)-\exp\left(-\frac{(\gamma-(1-m-yb))^2}{2\Delta q}\right)\right]\\
+ \mathbb{E}_y \left[(1-m-yb)\left(1-Q\left(\frac{1-m-yb}{\sqrt{\Delta q}}\right)-Q\left(\frac{\gamma-(1-m-yb)}{\sqrt{\Delta q}}\right)\right)+\gamma Q\left(\frac{\gamma-(1-m-yb)}{\sqrt{\Delta q}}\right)\right],
\end{split}
\end{equation}
\begin{equation}
\begin{split}
K_q=\sqrt{\frac{\Delta q}{2\pi}}\mathbb{E}_y\left[(1-m-yb)\exp\left(-\frac{(1-m-yb)^2}{2\Delta q}\right)-(\gamma+1-m-yb)\exp\left(-\frac{(\gamma-(1-m-yb))^2}{2\Delta q}\right)\right]\\
+\mathbb{E}_y\left[\left(\Delta q + (1-m-yb)^2\right) \left(1-Q\left(\frac{1-m-yb}{\sqrt{\Delta q}}\right)-Q\left(\frac{\gamma - (1-m-yb)}{\sqrt{\Delta q}}\right)\right)+\gamma^2 Q\left(\frac{\gamma - (1-m-yb)}{\sqrt{\Delta q}}\right)\right]
.\end{split}
\end{equation}
The equation to determine the bias is 
\begin{equation}
\begin{split}
    \sqrt{\frac{\Delta q}{2\pi}}\mathbb{E}_y\left[y\exp\left(-\frac{(1-m-yb)^2}{2\Delta q}\right)-y\exp\left(-\frac{(\gamma-(1-m-yb))^2}{2\Delta q}\right)\right]+\gamma \mathbb{E}_y\left[yQ\left(\frac{\gamma-(1-m-yb)}{\sqrt{\Delta q}}\right)\right]
    \\+\mathbb{E}_y\left[y(1-m-yb)\left(1-Q\left(\frac{1-m-yb}{\sqrt{\Delta q}}\right)-Q\left(\frac{\gamma-(1-m-yb)}{\sqrt{\Delta q}}\right)\right)\right]=0.
    \end{split}
\end{equation}

\section{Bayes-optimality at $\lambda=\infty$, for $\rho=\tfrac{1}{2}$}
\label{app:BO_large_lambda}
In this section we will show how the result on Bayes-optimality for balanced clusters at large regularization arises.
First we start by considering the square loss. At $\rho=1/2$, it is straightforward to check from \eref{11} that $b=0$ and the generalization error, given by \eref{generalization} in the main text, is
\begin{equation}
    \varepsilon_{\rm gen}=Q\left(\frac{m}{\sqrt{\Delta q}}\right),
\end{equation}
where $m$ and $q$ are given by \eref{l2_m}-\eref{l2_q}, evaluated at $\rho=\tfrac{1}{2}$.
The Bayes-optimal error for this problem is given by \eref{Bayes} in the main text and reads
\begin{equation}
    \varepsilon_{\rm gen}^{\rm BO}=Q\left(\sqrt{\frac{\alpha}{\Delta(\Delta+\alpha)}}\right).
\end{equation}
Therefore, in order to reach Bayes-optimality, we need a weight vector $\bf w$ with an overlap $m$ and a length $q$ such that
\begin{equation}
    \sqrt{\frac{\alpha}{(\Delta+\alpha)}}=\frac{m}{\sqrt{ q}}=\left(\sqrt{\frac{\hat{q}}{\hat{m}^2}+1}\right)^{-1}. \label{eq:opt_condition}
\end{equation}
By using \eref{l2_mhat}-\eref{l2_qhat} evaluated at $\rho=\tfrac{1}{2}$, \eref{opt_condition} can be rewritten as
\begin{equation}
    \frac{\Delta q}{(1-m)^2}=0.\label{eq:opt_cond2}
\end{equation}
Eq. \eref{opt_cond2} is verified by the fixed point equations only at $\lambda \to \infty$. Indeed in this limit we find that
$$\gamma = \frac {\Delta}{\lambda} + o\left(\lambda^{-1}\right),$$
hence
$$ m = \frac {\alpha}{\lambda} + o\left(\lambda^{-1}\right)$$
and
$$ q = \frac {\alpha}{\lambda^2}\left(\Delta+\alpha \right) + o(\lambda^{-2}),$$ 
so that
$$
\frac m{\sqrt q} \to \sqrt{\frac{\alpha}{(\Delta+\alpha)}}.
$$
Therefore, as $\lambda$ grows and while the $\ell_2$ norm of the vector goes to zero, the vector aligns itself optimally to the hidden one and the generalization error becomes optimal.

%%%%%%
\begin{figure}
\includegraphics[scale=0.3]{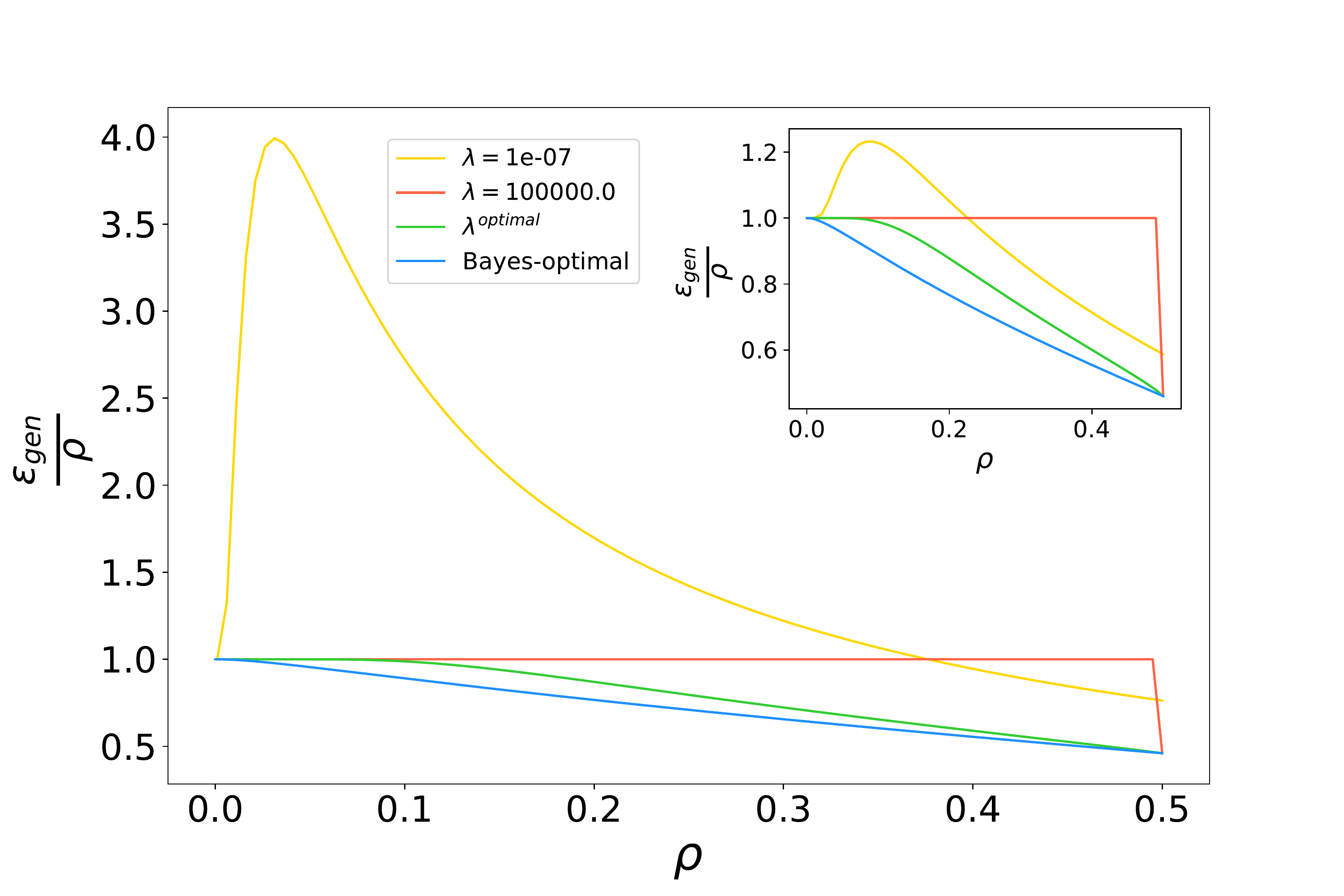}
\hspace{-0.6cm}
\includegraphics[scale=0.3]{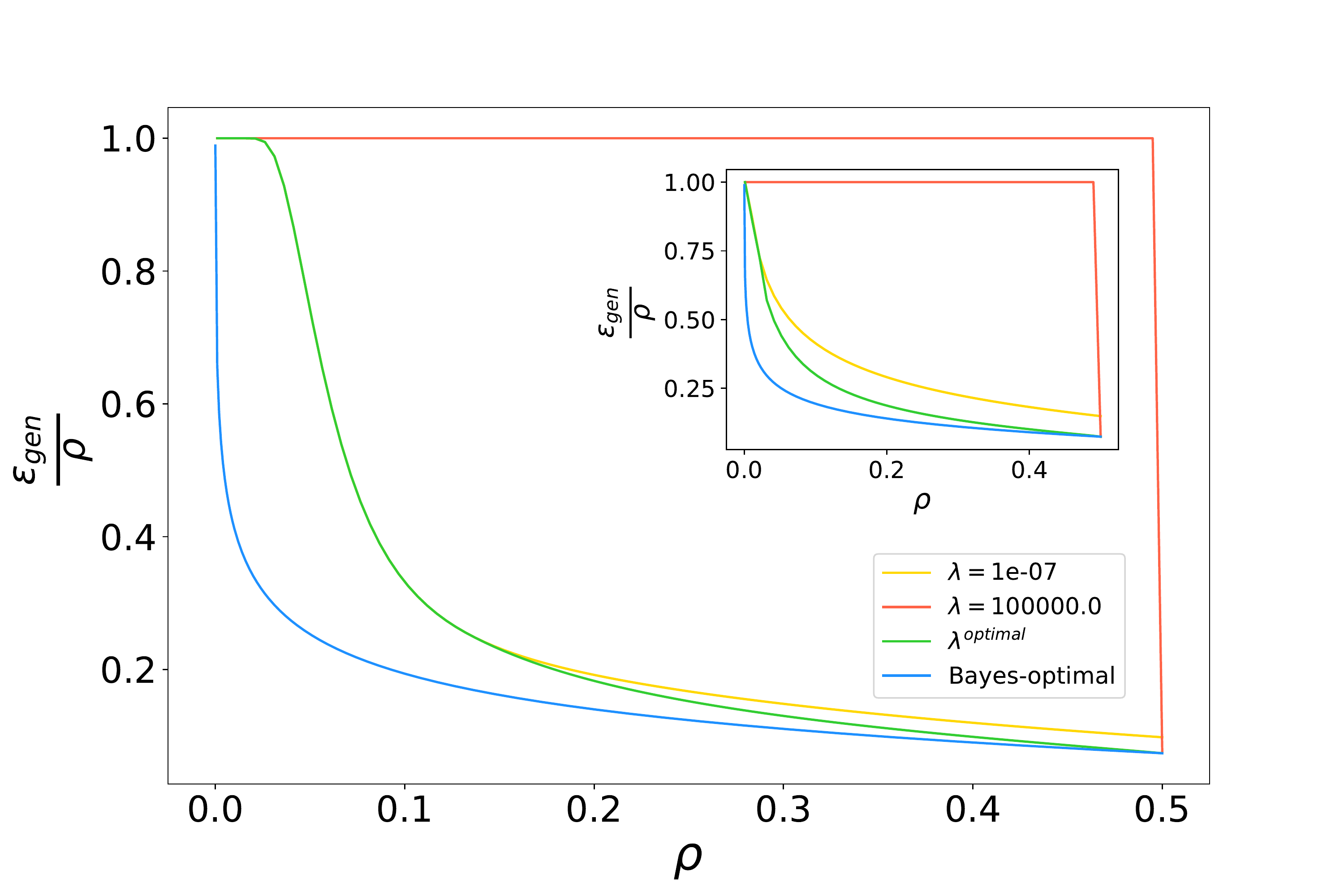}
\caption{ Generalization error as a function of $\rho$, at fixed $\alpha=1.2$ and $\Delta=1$ (left) and $\alpha=7$ and $\Delta=0.3$ (right), for the square loss compared to the Bayes-optimal performance. In the inset, the same figure for the hinge loss. The vertical axis is rescaled by $\rho$ for convenience. The error is computed at low ($\lambda=10^{-7}$), high ($\lambda=10^{5}$) and optimal regularization. We observe that Bayes-optimality at infinite regularization holds strictly at $\rho=1/2$.}
\label{gen_vs_rho_varying_lambda}
\end{figure}

It is then easy to see why this remains correct for any differentiable loss: as long as the $\ell_2$ norm vanishes when $\lambda \to \infty$, then one can expand 
$$
\ell(\bf w ^\top \bf x) = \ell(0) + \bf w ^\top \bf x \ell'(0) + o(q)
$$
so that any loss will behave like the square one. This is the origin of the peculiar behavior of Bayes optimally observed at $\lambda \to \infty$ for the symmetric case $\rho=1/2$. 
We observed numerically that this result is not valid anymore as soon as $\rho\neq 1/2$. This peculiar behaviour is shown in Fig.~\ref{gen_vs_rho_varying_lambda}, which depicts the generalization error, computed from the solution of \eref{1}-\eref{11} in the main text, as a function of $\rho$ at zero, infinite and optimal regularization for the square and hinge losses.

\section{Details on the numerics}
\label{app:numerics}
\subsection{Iteration of the fixed point equations}
%\textcolor{orange}{Maybe this subsection is not needed\\}
The solution $(q,m,b,\gamma)$ of the fixed point equations \eref{1}-\eref{6} can be obtained analytically only in the case of square loss. For the hinge and logistic loss, the equations have to be iterated until convergence. In our codes we used initialization $(q^{t=0},\gamma^{t=0},m^{t=0},b^{t=0})=(0.5,0.5,0.01,0)$. The stopping criterion for convergence consists in checking if the values of the generalization error at two consecutive iterations differ less than a threshold $eps$. In all figures, we used $eps\leq 10^{-8}$. 
\subsection{Simulations}
In order to check the validity of the fixed point equations \eref{1}-\eref{6} we computed numerically the solution of the optimization problem defined in \eref{loss}, and we averaged over multiple realizations of the noise. In the case of square loss, the solution is simply
\begin{equation}
    {\bf w}^{\rm square}=\left({\bf X}^\top {\bf X}+\lambda {\bf I}_d\right)^{-1} {\bf X}^\top {\bf y}.
\end{equation}
In the case of logistic and hinge loss, the solution can be computed by a standard gradient descent algorithm. In particular, in Fig.~\ref{fig:unreg_gen_vs_alpha} we used the Logistic Regression classifier provided by the scikitlearn package $linear\_model$ \cite{scikit}. In particular, we used the ``lbfgs" solver, with L2-penalty, tolerance $tol=10^{-5}$ for the stopping criterion and maximum number of iterations $max\_iter=10^{-5}$.
It is important to remind that all our analytic results are computed in the infinite-dimensional limit $d,n\rightarrow \infty$, while the ratio $\alpha=n/d$ remains finite. Therefore, all the simulations involve errors due to finite size effects. However, we found a very good agreement bewteen theory and simulations already at relatively small dimensionality ($ d\leq5000 $). The only case in which finite size effects prevent simulations to match our theoretical predictions is the behaviour of the generalization error at large regularization $\lambda$, at $\rho = 1/2$. Since at all finite dimensions $d$ the effective clusters size is $\rho\neq 1/2$, the result of reaching Bayes-optimality at $\lambda\rightarrow\infty$ cannot be obtained in simulations, since it holds strictly at $\rho=1/2$. However, we obtain greater and greater precision, i.e. the minimum of the generalization error moving towards higher values of $\lambda$ (see Fig. \ref{fig:gen_vs_L_L2}), as $d$ increases.

%\end{document}

% This document was modified from the file originally made available by
% Pat Langley and Andrea Danyluk for ICML-2K. This version was created
% by Iain Murray in 2018, and modified by Alexandre Bouchard in
% 2020. Previous contributors include Dan Roy, Lise Getoor and Tobias
% Scheffer, which was slightly modified from the 2010 version by
% Thorsten Joachims & Johannes Fuernkranz, slightly modified from the
% 2009 version by Kiri Wagstaff and Sam Roweis's 2008 version, which is
% slightly modified from Prasad Tadepalli's 2007 version which is a
% lightly changed version of the previous year's version by Andrew
% Moore, which was in turn edited from those of Kristian Kersting and
% Codrina Lauth. Alex Smola contributed to the algorithmic style files.

\end{document}